\documentclass[10pt,twocolumn,letterpaper]{article}

\usepackage{cvpr}
\usepackage{times}
\usepackage{epsfig}
\usepackage{graphicx}
\usepackage{amsmath}
\usepackage{amssymb}
\usepackage[title]{appendix}
\usepackage{mathrsfs}
\usepackage[ruled]{algorithm2e}
\usepackage{booktabs}
\usepackage{multirow}
\usepackage{subfigure}
\usepackage{rotating}
\usepackage[noend]{algpseudocode}
\usepackage[utf8x]{inputenc}
\DeclareMathOperator*{\argmax}{arg\,max}

\newcommand{\R}{\mathbb{R}}
\usepackage[breaklinks=true,bookmarks=false]{hyperref}

\cvprfinalcopy % *** Uncomment this line for the final submission

\usepackage{xcolor}

\newcommand{\p}[1]{\textbf{#1}:}

\def\cvprPaperID{2209} % *** Enter the CVPR Paper ID here

\begin{document}

%%%%%%%%% TITLE
\title{Learning to Learn Single Domain Generalization}

%\author{Fengchun Qiao$^1$ \quad Long Zhao$^2$ \quad Xi Peng$^1$ \\
%$^1$University of Delaware \quad $^2$Rutgers University\\
%{\tt\small \{fengchun,xipeng\}@udel.edu, lz311@cs.rutgers.edu}
%}

\author{Fengchun Qiao\\
University of Delaware\\
{\tt\small fengchun@udel.edu}
\and
Long Zhao\\
Rutgers University\\
{\tt\small lz311@cs.rutgers.edu}
\and
Xi Peng\\
University of Delaware\\
{\tt\small xipeng@udel.edu}
}

\maketitle
%\thispagestyle{empty}

%%%%%%%%% ABSTRACT
\begin{abstract}

We are concerned with a worst-case scenario in model generalization, in the sense that a model aims to perform well on many unseen domains while there is only one single domain available for training. We propose a new method named adversarial domain augmentation to solve this Out-of-Distribution (OOD) generalization problem. The key idea is to leverage adversarial training to create ``fictitious'' yet ``challenging'' populations, from which a model can learn to generalize with theoretical guarantees. To facilitate fast and desirable domain augmentation, we cast the model training in a meta-learning scheme and use a Wasserstein Auto-Encoder (WAE) to relax the widely used worst-case constraint. Detailed theoretical analysis is provided to testify our formulation, while extensive experiments on multiple benchmark datasets indicate its superior performance in tackling single domain generalization. \footnotetext[1]{The source code and pre-trained models are publicly available at: \url{https://github.com/joffery/M-ADA}.}

\end{abstract}
%\vspace{-1em}

%%%%%%%%% BODY TEXT
\section{Introduction}
Recent years have witnessed rapid deployment of machine learning models for broad applications~\cite{lecun2015deep,shrivastava2017learning,konstantinos2018using,zhao2019semantic}. A key assumption underlying the remarkable success is that the training and test data usually follow similar statistics. Otherwise, even strong models ({\it e.g.,} deep neural networks) may break down on unseen or Out-of-Distribution (OOD) test domains~\cite{balaji2018metareg}. Incorporating data from multiple training domains somehow alleviates this issue~\cite{li2017deeper}, however, this may not always be applicable due to data acquiring budget or privacy issue. An interesting yet seldom investigated problem then arises: Can a model generalize from one source domain to many unseen target domains? In other words, how to maximize the model generalization when there is only a single domain available for training?

\begin{figure}[!t]
\centering
\includegraphics[width=0.46\textwidth]{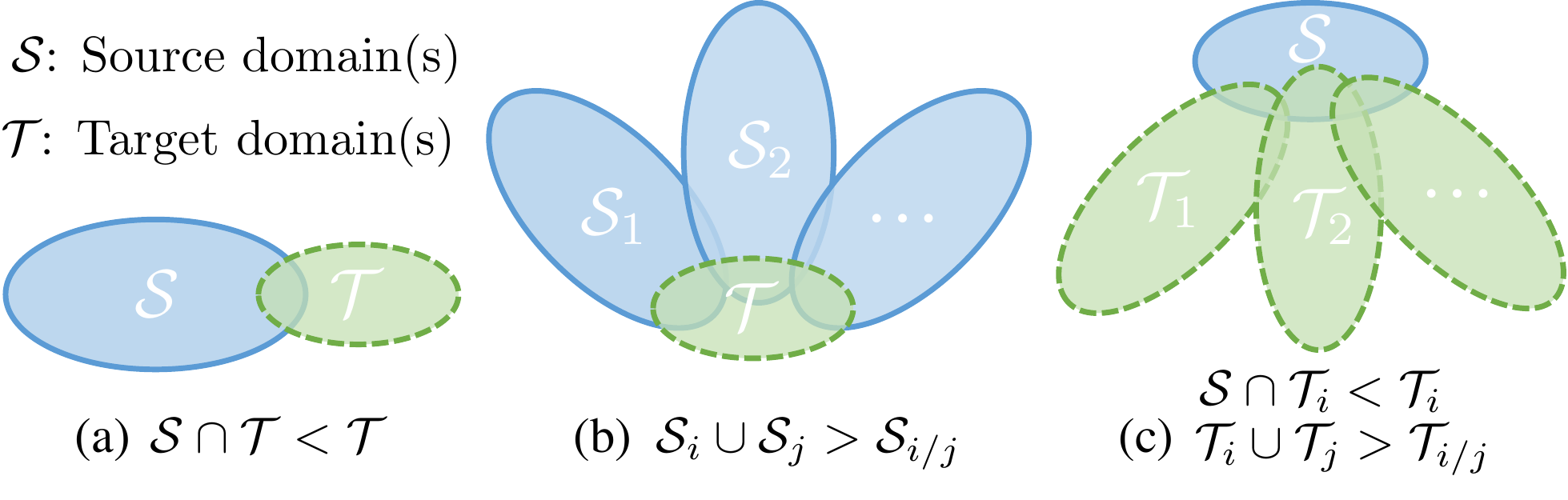}
\caption{The domain discrepancy: (a) domain adaptation, (b) domain generalization, and (c) single domain generalization.}
\label{fig:problem_illustration}
\end{figure}

The discrepancy between source and target domains, also known as domain or covariate variant~\cite{storkey2006mixture}, has been intensively studied in {\it domain adaptation}~\cite{motiian2017few,murez2018image,xu2019dsne,liu2019transferable} and {\it domain generalization}~\cite{muandet2013domain,ghifary2015domain,li2018learning,carlucci2019jigasaw}. Despite of their various success in tackling ordinary domain discrepancy issue, however, we argue that existing methods can hardly succeed in the aforementioned single domain generalization problem. As illustrated in Fig.~\ref{fig:problem_illustration}, the former usually expects the availability of target domain data (either labeled or unlabeled); While the latter, on the other hand, always assumes multiple (rather than one) domains are available for training. This fact emphasizes the necessity to develop a new learning paradigm for {\it single domain generalization}.

In this paper, we propose {\it adversarial domain augmentation} (Sec.~\ref{sec:3.1}) to solve this challenging task. Inspired by the recent success of adversarial training~\cite{peng2018jointly,tang2019adaptive,szegedy2014intriguing,ratner2017learning,liu2019transferable}, we cast the single domain generalization problem in a worst-case formulation~\cite{sinha2017certifying,lee2018minimax}. The goal is to use single source domain to generate ``fictitious'' yet ``challenging'' populations, from which a model can learn to generalize with theoretical guarantees (Sec.~\ref{sec:4}).

However, technical barriers exist when applying adversarial training for domain augmentation. On the one hand, it is hard to create ``fictitious'' domains that are largely different from the source, due to the contradiction of semantic consistency constraint~\cite{goodfellow2014adv} in worst-case formulation. On the other hand, we expect to explore many ``fictitious'' domains to guarantee sufficient coverage, which may result in significant computational overhead. To circumvent these barriers, we propose to {\it relax the worst-case constraint} (Sec.~\ref{sec:3.2}) via a Wasserstein Auto-Encoder (WAE)~\cite{tolstikhin2018wasserstein} to encourage large domain transportation in the input space. Moreover, rather than learning a series of ensemble models~\cite{volpi2018generalizing}, we organize adversarial domain augmentation via {\it meta-learning}~\cite{finn2017model} (Sec.~\ref{sec:3.3}), yielding a highly efficient model with improved single domain generalization.

The primary contribution of this work is a meta-learning based scheme that enables single domain generalization, an important yet seldom studied problem. We achieve the goal by proposing adversarial domain augmentation, while at the same time, relaxing the widely used worst-case constraint. We also provide detailed theoretical understanding to testify our solution. Extensive experiments indicate that our method marginally outperforms state of the art in single domain generalization of benchmark datasets including {\it Digits}, {\it CIFAR-10-C}~\cite{hendrycks2019benchmarking}, and {\it SYTHIA}~\cite{ros2016synthia}.

%However, technical barriers exist when applying adversarial training for domain augmentation. Firstly, it is hard to generate ``ficitious'' domains that are largely different from source domain due to the worst-case constraint. To address this limitation, we propose to {\it relax Wasserstein distance constraint}(Section~\ref{sec:3.2}), which merely enforce semantic consistency in embedding space, to encourage domain transportation in input space. Secondly, exploring many ``fictitious'' domains may result in significant computational overhead. Rather than learning a series of ensemble models[], we organize adversarial domain augmentation in {\it a learning to learn framework} (Section~\ref{sec:3.3}), yielding a single model with improved single domain generalization.

\section{Related Work}

\p{Domain discrepancy}
%\textcolor{blue}{We should put 2 paragraphs here: the 1st paragraph talks about domain shift issue and domain adaptation; The 2nd paragraph talks about domain generalization. (You can put ERM and GUD discussion here.)}
Domain discrepancy brought by domain or covariance shifts~\cite{storkey2006mixture} severely degrades the model performance on cross-domain recognition. %~\cite{peng2017recons}.
The models trained using Empirical Risk Minimization~\cite{koltchinskii2011oracle} usually perform poorly on unseen domains. To reduce the discrepancy across domains, a series of methods are proposed for unsupervised~\cite{murez2018image,shu2018dirt,french2017self,russo2018source,sankaranarayanan2018generate} or supervised domain adaptation~\cite{motiian2017unified,xu2019dsne}.
%{\color{blue} Specially, Sun \etal~\cite{sun2019test} proposed test-time training for OOD generalization, where single unlabeled test instance is trained through self-supervised learning.}
Some recent work also focused on few-shot domain adaptation~\cite{motiian2017few} where only a few labeled samples from target domain are involved in training.

Different from domain adaptation, domain generalization aims to learn from multiple source domains without any access to target domains.
Most previous methods either tried to learn a domain-invariant space to align domains~\cite{muandet2013domain,ghifary2015domain,grubinger2017multi,li2017deeper,zhao2020knowledge} or aggregate domain-specific modules~\cite{mancini2018robust,mancini2018best}. Recently, Carlucci \etal~\cite{carlucci2019jigasaw} solved this problem by jointly learning from supervised and unsupervised signals from images. In data level, gradient-based domain perturbation~\cite{shankar2018generalizing} and adversarial training methods~\cite{volpi2018generalizing} are proposed to improve generalization. In particular, \cite{volpi2018generalizing} is designed for single domain generalization and achieves better performance through an ensemble model. Compared to \cite{volpi2018generalizing}, we aim at creating large domain transportation for ``fictitious'' domains and devising a more efficient meta-learning scheme within a single unified model.

%With the purpose of directly generalizing to unseen target domains with a single model, domain generalization has been actively studied in recent years, where models are trained on multiple source domains without access to any target domain. Existing methods can be divide into four categories: learning a domain-invariant space to align different source domains $\mathcal{D}_s$ \cite{muandet2013domain,ghifary2015domain,grubinger2017multi,li2017deeper}, aggregating multiple domain-specific modules \cite{mancini2018robust,mancini2018best}, training with auxiliary losses to regularize models \cite{balaji2018metareg,carlucci2019jigasaw}, generating more data to better overlap with target domains $\mathcal{D}_t$ \cite{volpi2018generalizing,shankar2018generalizing}. 
\p{Adversarial training} Adversarial training~\cite{goodfellow2014adv} is proposed for improving model robustness against adversarial perturbations or attacks. Madry \etal~\cite{madry2017pgd} provided evidence that deep neural networks is capable of resistant to adversarial attacks through reliable adversarial training methods. Further, Sinha \etal~\cite{sinha2017certifying} proposed principled adversarial training through the lens of distributionally robust optimization. More recently, Stutz \etal~\cite{stutz2019disentangling} pointed out that on-manifold adversarial training boosts generalization, and hence models with both robustness and generalization can be obtained at the same time. Peng \etal~\cite{peng2018jointly} proposed to learn robust models via perturbed examples. In our work, we generate ``fictitious'' domains through adversarial training to improve single domain generalization.

\p{Meta-learning}
%\textcolor{blue}{Talk about meta-learning background and various applications. Especially, we should mention use meta-learning for few-shot learning.}
Meta-learning~\cite{schmidhuber1987evolutionary,thrun2012learning} is a long standing topic in how to learn new concepts or tasks fast with a few training examples. It has been widely used in optimization of deep neural networks~\cite{andrychowicz2016learning,lilearning} and few-shot classification~\cite{koch2015siamese,vinyals2016matching,snell2017prototypical}.
Recently, Finn \etal~\cite{finn2017model} proposed a Model-Agnostic Meta-Learning (MAML) procedure for few-shot learning and reinforcement learning.
The objective of MAML is to find a good initialization which can be fast adapted to new tasks within few gradient steps. Li \etal~\cite{li2018learning} proposed a MAML-based approach to solve domain generalization.
%{\color{blue} In \cite{li2018learning}, meta-test domain is randomly selected from source domains, while ``fictitious''  domains are trained in meta-test of our method. }
Balaji \etal~\cite{balaji2018metareg} proposed to learn an adaptive regularizer through meta-learning for cross-domain recognition. However, neither of them is applicable for single domain generalization.
Instead, in this paper, we propose a MAML-based meta-learning scheme to efficiently train models on ``fictitious'' domains for single domain generalization. We show that the learned model is robust to unseen target domains while it can also be easily leveraged for few-shot domain adaptation. 

\section{Method}

%We aim at solving the problem of single domain generalization, where the model is trained with $(\mathbf{x}, \mathbf{y})$ from only one source domain $\mathcal{S}$ and expected to generalize well on unseen target domains $\mathcal{T}$. 
%Domain generalization and domain adaptation are two kinds methods to address distribution shifts in cross-domain recognition. Multiple labeled source domains are required in domain generalization and partial or full target domains are involved in the training of domain adaptation. As a result, neither of them can be applied in single domain generalization where only one source domain is acquirable and target domains are unseen.
%From the perspective of adversarial training \cite{goodfellow2014adv}, a sample from $\mathcal{T}$ can be viewed as the perturbation against the sample from $\mathcal{S}$. Therefore, the goal of adversarial training in this paper is to devise the samples involved during training procedures, so that the robustness of the model against possible $\mathcal{T}$ can be improved.
% Inspired by recent achievements of robust optimization and adversarial training \cite{sinha2017certifying}, we consider the worst-case problem that is distance $\rho$ away from the source domain $\mathcal{S}$:

We aim at solving the problem of single domain generalization: A model is trained on only one source domain $\mathcal{S}$ but is expected to generalize well on many unseen target domains $\mathcal{T}$. A promising solution of this challenging problem, inspired by many recent achievements~\cite{ratner2017learning,volpi2018generalizing,liu2019transferable}, is to leverage adversarial training~\cite{goodfellow2014adv,szegedy2014intriguing}. The key idea is to learn a robust model that is resistant to out-of-distribution perturbations.
More specifically, we can learn the model by solving a worst-case problem~\cite{sinha2017certifying}: 
\begin{equation}
\underset{\theta}{\operatorname{min}} \sup _{\mathcal{T}: D\left(\mathcal{S},\mathcal{T}\right) \leq \rho} \mathbb{E}[\mathcal{L}_{\mathrm{task}}(\theta ;\mathcal{T})],
\label{eq:worst}
\end{equation}
%where $D$ is a distance metric to measure the distribution distance between two domains, and $\theta$ are the model parameters optimized according to a task-specific objective function $\mathcal{L}_{\mathrm{task}}$. Here, we focus on classification problems where $\mathcal{L}_{\mathrm{task}}$ is defined as the cross-entropy loss:
where $D$ is a similarity metric to measure the domain distance and $\rho$ denotes the largest domain discrepancy between $\mathcal{S}$ and $\mathcal{T}$. $\theta$ are model parameters that are optimized according to a task-specific objective function $\mathcal{L}_{\mathrm{task}}$. Here, we focus on classification problems using cross-entropy loss:
\begin{equation}\label{eq:ce}
\mathcal{L}_{\mathrm{task}}(\mathbf{y}, \hat{\mathbf{y}}) = -\sum_i y_i \log (\hat{y}_{i}),
\end{equation}
where $\hat{\mathbf{y}}$ is \textit{softmax} output of the model; $\mathbf{y}$ is the one-hot vector representing the ground truth class; $y_i$ and $\hat{y}_{i}$ represent the $i$-th dimension of $\mathbf{y}$ and $\hat{\mathbf{y}}$, respectively.

%In this paper, we propose a new method named \textit{Meta-Learning based Adversarial Domain Augmentation} (M-ADA) for single domain generalization with the worst-case guarantee. To be specific, in order to cover as many unseen target domains $\mathcal{T}$ as possible during training procedures, ``fictitious'' domains $\mathcal{S}^+$ are augmented from the source domain $\mathcal{S}$ by relaxing the semantic consistency constraint in Eq.~\eqref{eq:worst}. The semantic consistency constraint and distributional relaxation are presented in Secs.~\ref{sec:3.1} and \ref{sec:3.2}, respectively. Then we introduce a meta-learning scheme in Sec.~\ref{sec:3.3} to organize augmented domains $\mathcal{S}^+$ for efficient training. In Sec.~\ref{sec:4}, a theoretical analysis is provided to further testify our solution. We illustrate our idea in Fig.~\ref{fig_model}.

%The adversarial training performs a \textit{minimax} game: a \textit{maximization} phase where adversarial samples of $\mathbf{x}$ are generated by gradient ascent which increases $\mathcal{L}_{\mathrm{task}}$ and a \textit{minimization} phase where the robustness of the model is improved through training on these adversarial samples.

\begin{figure}[t]
\centering
\includegraphics[width=1.0\linewidth]{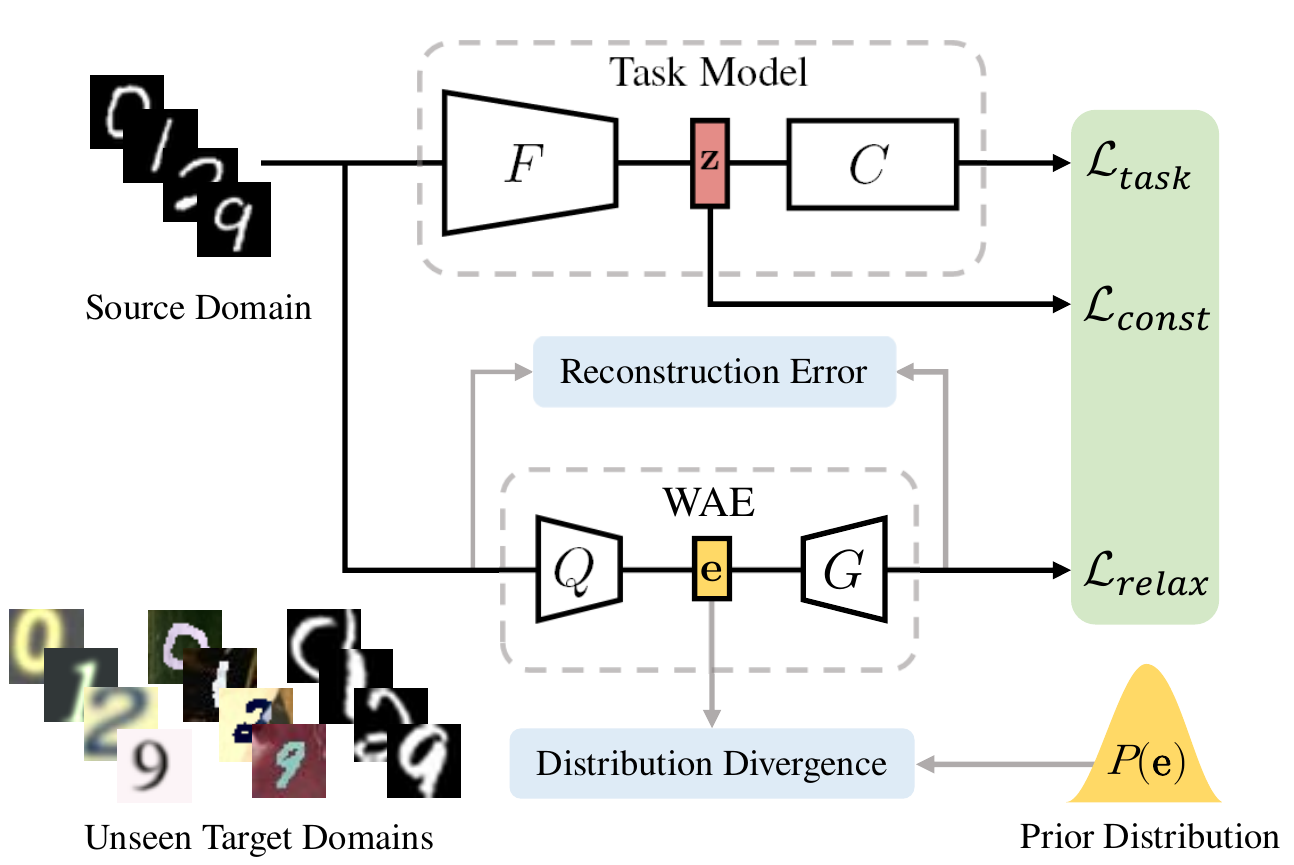}
\caption{Overview of adversarial domain augmentation.}
\label{fig_model}
\end{figure}

Following the worst-case formulation~\eqref{eq:worst}, we propose a new method, \textit{Meta-Learning based Adversarial Domain Augmentation} (M-ADA), for single domain generalization. Fig.~\ref{fig_model} presents an overview of our approach. We create ``fictitious'' yet ``challenging'' domains by leverage adversarial training to augment the source domain in Sec.~\ref{sec:3.1}. The task model learns from the domain augmentations with the assistance of a Wasserstein Auto-Encoder (WAE), which relaxes the worst-case constraint in Sec.~\ref{sec:3.2}. We organize the joint training of task model and WAE, as well as the domain augmentation procedure, in a learning to learn framework as described in Sec.~\ref{sec:3.3}. Finally, we present theoretical analysis to prove the worst-case guarantee in Sec.~\ref{sec:4}.

% Typically, to guarantee sufficient coverage of unseen target domains $\mathcal{T}$, we expect to explore a large number of augmented domains $\mathcal{S}^+$. On the one hand, there is no guarantee that each augmented domain is distributionally different from the source domain $\mathcal{S}$. 
%However, $\mathcal{L}_{\mathrm{const}}$ yields limited domain transportation since it severely constrains the semantic distance between the samples and their perturbations. 
%Hence, $\mathcal{L}_{\mathrm{relax}}$ is proposed to relax the semantic consistency constraint and create large domain transportation. 
% To tackle this issue,  $\mathcal{L}_{\mathrm{relax}}$ is proposed to relax the semantic consistency constraint and create large domain transportation. 
%The implementation of $\mathcal{L}_{\mathrm{relax}}$ is discussed in Sec.~\ref{sec:3.2}. 
% In Sec.~\ref{sec:4}, we proof that maximizing $\mathcal{L}_{\mathrm{relax}}$ is the direct derivation of a relaxed version of Eq.~\eqref{eq:worst}.

% On the other hand, exploring many augmented domains would result in significant computational overhead. For example, previous work \cite{volpi2018generalizing} has to learn a series of ensemble models to generalize from a single domain. To organize $\mathcal{S}^+$ for efficient training in a single unified model, a Meta-Learning training scheme is proposed for fast adaptation from $\mathcal{S}$ to $\mathcal{S}^+$, and further improve generalization learning, which will be discussed in Sec.~\ref{sec:3.3}.

\subsection{Adversarial Domain Augmentation} \label{sec:3.1}
% Models trained only on the source domain may generalize poorly on unseen domains. 
Our goal is to create multiple augmented domains from the source domain. Augmented domains are required to be distributionally different from the source domain so as to mimic unseen domains. In addition, to avoid divergence of augmented domains, 
the worst-case guarantee defined in Eq.~\eqref{eq:worst} should also be satisfied.

To achieve this goal, we propose Adversarial Domain Augmentation. Our model consists of a task model and a WAE shown in Fig.~\ref{fig_model}. In Fig.~\ref{fig_model}, the task model consists of a feature extractor $F: \mathcal{X} \to \mathcal{Z}$ mapping images from input space to embedding space, and a classifier $C: \mathcal{Z} \to \mathcal{Y}$ used to predict labels from embedding space. Let $\mathbf{z}$ denote the latent representation of $\mathbf{x}$ which is obtained by $\mathbf{z} = F(\mathbf{x})$. The overall loss function is formulated as follows: 
\begin{equation}\label{eq:ada}
\mathcal{L}_\text{ADA} = \underbrace{\mathcal{L}_{\mathrm{task}}(\theta;\mathbf{x})}_{\mathrm{Classification}}  - \alpha \underbrace{\mathcal{L}_{\mathrm{const}}(\theta;\mathbf{z})}_{\mathrm{Constraint}}+\beta\underbrace{ \mathcal{L}_{\mathrm{relax}}(\psi;\mathbf{x})}_{\mathrm{Relaxation}}, 
\end{equation}
where $\mathcal{L}_{\mathrm{task}}$ is the classification loss defined in Eq.~\eqref{eq:ce}, $\mathcal{L}_{\mathrm{const}}$ is the worst-case guarantee defined in Eq.~\eqref{eq:worst}, and $\mathcal{L}_{\mathrm{relax}}$ guarantees large domain transportation defined in Eq.~\eqref{eq:relax}. $\psi$ are parameters of the WAE.
$\alpha$ and $\beta$ are two hyper-parameter to balance $\mathcal{L}_{\mathrm{const}}$ and $\mathcal{L}_{\mathrm{relax}}$. 

Given the objective function $\mathcal{L}_\text{ADA}$, we employ an iterative way to generate the adversarial samples $\mathbf{x}^+$ in the augmented domain $\mathcal{S}^+$:
\begin{equation}
\mathbf{x}^+_{t+1} \gets \mathbf{x}^+_{t} +  \gamma\nabla_{\mathbf{x}^+_{t}}\mathcal{L}_\text{ADA}(\theta,\psi;\mathbf{x}^+_{t},\mathbf{z}^+_{t}),
\label{eq:ascent}
\end{equation}
where $\gamma$ is the learning rate of gradient ascent. A small number of iterations are required to produce sufficient perturbations and create desirable adversarial samples.

$\mathcal{L}_{\mathrm{const}}$ imposes semantic consistency constraint to adversarial samples so that $\mathcal{S}^+$ satisfies $D\left(\mathcal{S},\mathcal{S}^+\right)\leq \rho$.
More specifically, we follow~\cite{volpi2018generalizing} to measure the Wasserstein distance between $\mathcal{S}^+$ and $\mathcal{S}$ in the embedding space:
\begin{equation}\label{eq:constraint}
\mathcal{L}_{\mathrm{const}}= \frac{1}{2}\Vert \mathbf{z}-\mathbf{z}^+\Vert_{2}^{2}+\infty \cdot \mathbf{1}\left\{\mathbf{y} \neq \mathbf{y}^+\right\},
\end{equation}
where $\mathbf{1}\{\cdot\}$ is the 0-1 indicator function and $\mathcal{L}_{\mathrm{const}}$ will be $\infty$ if the class label of $\mathbf{x}^+$ is different from $\mathbf{x}$. 
% {\color{blue} We assume $y^+ == y$ is always true given small enough step size, which simplifies implementation without compromising performance.} 
Intuitively, $\mathcal{L}_{\mathrm{const}}$ controls the ability of generalization outside the source domain measured by Wasserstein distance~\cite{villani2003topics}.
% In the conventional setting of adversarial training, the worst-case problem is handled by only $\mathcal{L}_{\mathrm{task}}$ and $\mathcal{L}_{\mathrm{const}}$. 
However, $\mathcal{L}_{\mathrm{const}}$ yields limited domain transportation since it severely constrains the semantic distance between the samples and their perturbations. 
Hence, $\mathcal{L}_{\mathrm{relax}}$ is proposed to relax the semantic consistency constraint and create large domain transportation. 
The implementation of $\mathcal{L}_{\mathrm{relax}}$ is discussed in Sec.~\ref{sec:3.2}.

\subsection{Relaxation of Wasserstein Distance Constraint}\label{sec:3.2}
\label{sec_wdis}

\begin{figure}[t]
\centering
\includegraphics[width=1.0\linewidth]{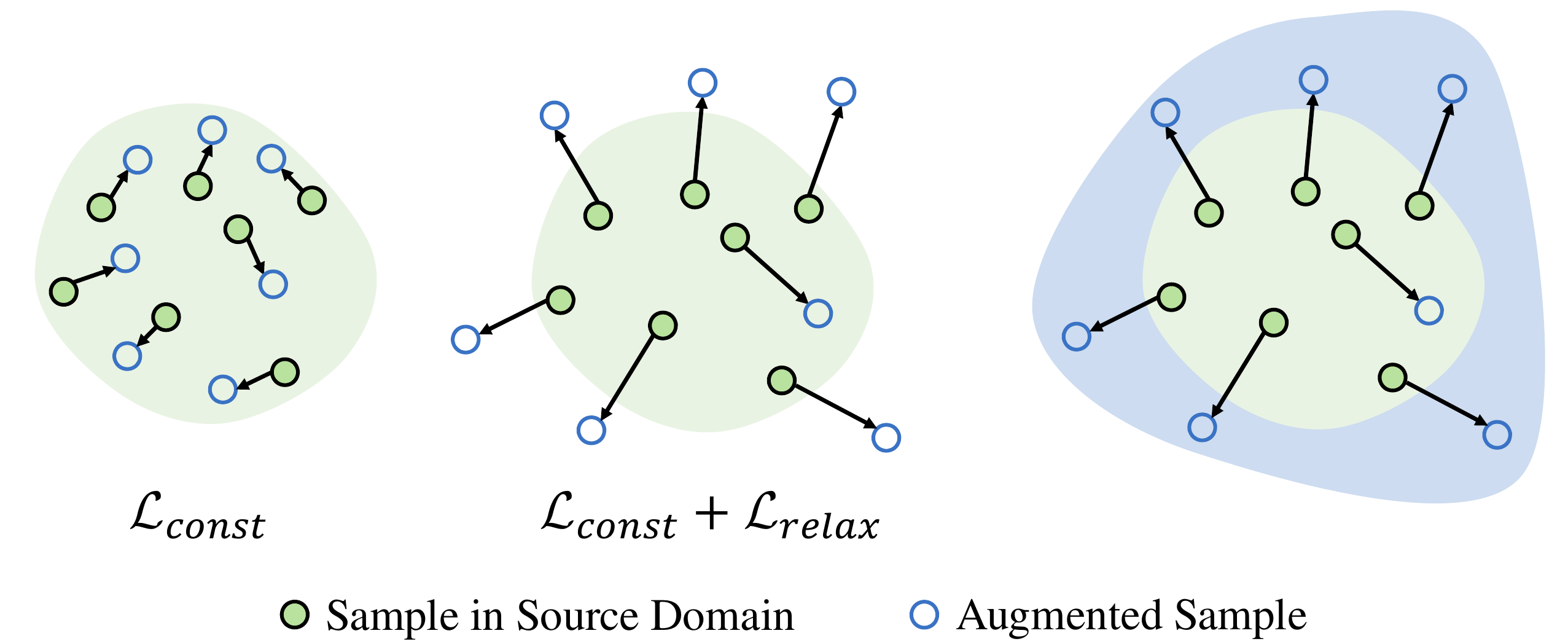}
\caption{Motivation of $\mathcal{L}_{\mathrm{relax}}$. {\bf Left:} The augmented samples may be close to the source domain if applying $\mathcal{L}_{\mathrm{const}}$. {\bf Middle:} We expect to create out-of-domain augmentations by incorporating $\mathcal{L}_{\mathrm{relax}}$. {\bf Right:} This would yield an enlarged training domain.}
\label{fig_ada}
\end{figure}

Intuitively, 
% in order to better mimic unseen target domains $\mathcal{T}$, 
we expect the augmented domains $\mathcal{S}^+$ are largely different from the source domain $\mathcal{S}$.
In other words, we want to maximize the domain discrepancy between $\mathcal{S}^+$ and $\mathcal{S}$.
However, the semantic consistency constraint $\mathcal{L}_{\mathrm{const}}$ would severely limits the domain transportation from $\mathcal{S}$ to $\mathcal{S}^+$, posing new challenges to generate desirable $\mathcal{S}^+$. To address this issue, we propose $\mathcal{L}_{\mathrm{relax}}$ to encourage out-of-domain augmentations.
% which effectively enlarge the training domain.
We illustrate the idea in Fig.~\ref{fig_ada}.

Specifically, we employ Wasserstein Auto-Encoders (WAEs)~\cite{tolstikhin2018wasserstein} to implement $\mathcal{L}_{\mathrm{relax}}$. Let $V$ denote the WAE parameterized by $\psi$. $V$ consists of an encoder $Q(\mathbf{e}|\mathbf{x})$ and a decoder $G(\mathbf{x}|\mathbf{e})$ where $\mathbf{x}$ and $\mathbf{e}$ denote inputs and bottleneck embedding, respectively. Additionally, we use a distance metric $\mathcal{D}_{\mathbf{e}}$ to measure the divergence between $Q(\mathbf{x})$ and a prior distribution $P(\mathbf{e})$, which can be implemented as either \textit{Maximum Mean Discrepancy} (MMD) or GANs \cite{goodfellow2014generative}. We can learn $V$ by optimizing:
\begin{equation}
\min _{\psi} [ \Vert G(Q(\mathbf{x}))- \mathbf{x} \Vert^2 + \lambda \mathcal{D}_{\mathbf{e}} (Q(\mathbf{x}),P(\mathbf{e}))],
\label{eq:wae}
\end{equation}
where $\lambda$ is a hyper-parameter. After pre-training $V$ on the source domain $S$ offline, we keep it frozen and maximize the reconstruction error $\mathcal{L}_{\mathrm{relax}}$ for domain augmentation:
\begin{equation}
\mathcal{L}_{\mathrm{relax}}= \Vert \mathbf{x}^+- V(\mathbf{x}^+) \Vert^2.
\label{eq:relax}
\end{equation}

Different from Vanilla or Variation Auto-Encoders \cite{kingma2013auto}, WAEs employ the Wasserstein metric to measure the distribution distance between the input and reconstruction. Hence, the pre-trained $V$ can better capture the distribution of the source domain and maximizing $\mathcal{L}_{\mathrm{relax}}$ creates large domain transportation. Comparison of different $\mathcal{L}_{\mathrm{relax}}$ is also provided in the supplementary.

In this work, $V$ acts as a \textit{one-class discriminator} to distinguish whether the augmentation is outside the source domain, which is significantly different from the traditional discriminator of GANs \cite{goodfellow2014generative}. And it is also different from the domain classifier widely used in domain adaptation \cite{liu2019transferable}, since there is only one source domain available. As a result, $\mathcal{L}_{\mathrm{relax}}$ together with $\mathcal{L}_{\mathrm{const}}$ are used to ``push away'' $\mathcal{S}^+$ in input space and ``pull back'' $\mathcal{S}^+$ in the embedding space simultaneously. In Sec.~\ref{sec:4}, we show that $\mathcal{L}_{\mathrm{relax}}$ and $\mathcal{L}_{\mathrm{const}}$ are derivations of two Wasserstein distance metrics defined in the input space and embedding space, respectively.

\subsection{Meta-Learning Single Domain Generalization}\label{sec:3.3}

% Existing methods \cite{goodfellow2014adv,madry2017pgd,sinha2017certifying} for adversarial training usually perform optimization on adversarial samples and clean data together. These methods may be suitable for a small number of adversarial samples with imperceptible perturbations. However, when the domain transportation and the number of augmented domains $\mathcal{S}^+$ are large enough, the distributional difference and efficiency of training cannot be overlooked. Because samples from $\mathcal{S}^+$ may make the training unstable and enlarge the variance of models. To address this problem, \cite{volpi2018generalizing} proposed  an ensemble method to aggregate prediction results of several models under different semantic consistency constraints. Instead, we aim at solving this problem in a single unified model.

% To make the classification model generalize to as many unseen target domains $\mathcal{T}$ as possible and improve the stability of training, a meta-learning training scheme is used to organize $\mathcal{S}^+$ for efficient training.

To efficiently organize the model training on the source domain $S$ and augmented domains $\mathcal{S}^+$, we leverage a meta-learning scheme to train a single model. To mimic real domain-shifts between the source domain $S$ and target domain $T$, at each learning iteration, we perform meta-train on the source domain $\mathcal{S}$ and meta-test on all augmented domains $\mathcal{S}^+$. Hence, after many iterations, the model is expected to achieve good generalization on the final target domain $\mathcal{T}$ during evaluation.

% In contrast to prior work~\cite{volpi2018generalizing}, which trains a series of ensemble model on both the source domain $\mathcal{S}$ and augmented domains $\mathcal{S}^+$, we cast the model training in a meta-learning scheme so that the learned model can be efficiently generalized to unseen target domains $\mathcal{T}$. To achieve this, at each learning iteration, we perform meta-train on the source domain $\mathcal{S}$ and meta-test on all augmented domains $\mathcal{S}^+$. This is to mimic real domain-shifts between the source domain and target domain so that over many iterations, we can train a model to achieve good generalization in the final target domain $\mathcal{T}$ during evaluation.

\begin{algorithm}[t]
	\caption{The proposed Meta-Learning based Adversarial Domain Augmentation (M-ADA).}
	\LinesNumbered
	\label{alg:overrall}
	\KwIn{Source domain $\mathcal{S}$; Pre-train WAE $V$ on $\mathcal{S}$; Number of augmented domains $K$}
	\KwOut{Learned model parameters $\theta$ }
	
	\For{$k=1,...,K$}{
		Generate $\mathcal{S}^+_{k}$ from $\mathcal{S} \cup \{\mathcal{S}^+_{i}\}^{k-1}_{i=1} $ using Eq.~\eqref{eq:ascent} \\
		Re-train $V$ with $\mathcal{S}^+_{k}$ \\
		\textbf{Meta-train}: Evaluate $\mathcal{L}_{\mathrm{task}}(\theta; \mathcal{S})$ \wrt $\mathcal{S}$ \\
		Compute $\hat{\theta}$ using Eq.~\eqref{eq:meta-train} \\
		\For{$i=1,...,k$}{
			\textbf{Meta-test}: Evaluate $\mathcal{L}_{\mathrm{task}}(\hat{\theta}; \mathcal{S}^+_i))$ \wrt $\mathcal{S}^+_i$ \\
		}
		\textbf{Meta-update}: Update $\theta$ using Eq.~\eqref{eq:meta-update}
	}
\end{algorithm}

Formally, the proposed Meta-Learning based Adversarial Domain Augmentation (M-ADA) approach consists of three parts in each iteration during the training procedure: meta-train, meta-test and meta-update. In meta-train, $\mathcal{L}_{\mathrm{task}}$ is computed on samples from the source domain $\mathcal{S}$, and the model parameters $\theta$ is updated via one or more gradient steps with a learning rate of $\eta$:
\begin{equation}\label{eq:meta-train}
\hat{\theta} \gets \theta - \eta \nabla_{\theta} \mathcal{L}_{\mathrm{task}}(\theta; \mathcal{S}).
\end{equation}
Then we compute $\mathcal{L}_{\mathrm{task}}(\hat{\theta}; \mathcal{S}^+_k)$ on each augmented domain $\mathcal{S}^+_k$ in meta-test. At last, in meta-update, we update $\theta$ by the gradients calculated from a combined loss where  meta-train and meta-test are optimised simultaneously:
\begin{equation}\label{eq:meta-update}
\theta \gets \theta - \eta\nabla_{\theta} [\mathcal{L}_{\mathrm{task}}(\theta; \mathcal{S})+ \sum^{K}_{k=1}\mathcal{L}_{\mathrm{task}}(\hat{\theta}; \mathcal{S}^+_k)],
\end{equation}
%where we combine the loss computed on $\mathcal{S}$ and $\mathcal{S}^+$ together to make sure the model also performs well on some target domains $\mathcal{T}$ nearby $\mathcal{S}$. 
where $K$ is the number of augmented domains. 

%The entire training algorithm is summarized in Alg.~\ref{alg:overrall}. The advantages of our scheme come in two folds. First, in contrast to prior work~\cite{volpi2018generalizing}, which trains a series of ensemble models on both the source domain $\mathcal{S}$ and augmented domains $\mathcal{S}^+$, we efficiently organize the training in a single model.
%We experimentally show that our method outperforms \cite{volpi2018generalizing} marginally in terms of memory, speed and accuracy. 
%Second, the learned model parameters are sensitive to changes in the domain shifts, and one or a small number of gradient steps on a new target domain will produce maximally effective behavior on that domain. This prepares the learned model for fast adaptation, which makes it easy to extend M-ADA to \textit{few-shot domain adaptation}.

The entire training pipeline is summarized in Alg.~\ref{alg:overrall}. Our method has following merits. First, in contrast to prior work~\cite{volpi2018generalizing} that learns a series of ensemble models, our method achieves a single model for efficiency. In Sec.~\ref{sec:sota}, we prove that M-ADA outperforms \cite{volpi2018generalizing} marginally in terms of memory, speed and accuracy. 
Second, the meta-learning scheme prepares the learned model for fast adaptation: One or a small number of gradient steps will produce improved behavior on a new target domain. This enables M-ADA for \textit{few-shot domain adaptation} as shown in Sec~\ref{sec:few}.

\section{Theoretical Understanding} \label{sec:4}
We provide a detailed theoretical analysis of the proposed Adversarial Domain Augmentation.
Specifically, we show that the overall loss function defined in Eq.~\eqref{eq:ada} is a direct derivation of a relaxed worst-case problem.
% relaxes the semantic consistency constraint $\mathcal{L}_{\mathrm{const}}$ using the distributional relaxation $\mathcal{L}_{\mathrm{relax}}$. We proof that Eq.~\eqref{eq:ada} is a direct derivation of Eq.~\eqref{eq:worst} with a relaxed worst-case guarantee.
% We assume that $\mathcal{S}^+$ is neighboring $\mathcal{S}$ in the embedding space $\mathcal{Z}$ of the classification model, but still has a large domain transportation in the input space $\mathcal{X}$. 

Let $c: \mathcal{Z} \times \mathcal{Z} \to \R_+ \cup \{\infty\}$ be the ``cost'' for an adversary to perturb $\mathbf{z}$ to $\mathbf{z}^+$ in the embedding space. Let $d: \mathcal{X} \times \mathcal{X} \to \R_+ \cup \{\infty\}$ be the ``cost'' for an adversary to perturb $\mathbf{x}$ to $\mathbf{x}^+$ in the input space. The Wasserstein distances between $\mathcal{S}$ and $\mathcal{S}^+$ can be formulated as:
$W_c(\mathcal{S}, \mathcal{S}^+):=\inf _{M_{\mathbf{z}} \in \Pi(\mathcal{S}, \mathcal{S}^+)} \mathbb{E}_{M_{\mathbf{z}}}\left[c\left(\mathbf{z}, \mathbf{z}^+\right)\right]$ and
$W_d(\mathcal{S}, \mathcal{S}^+):=\inf _{M_{\mathbf{x}} \in \Pi(\mathcal{S}, \mathcal{S}^+)} \mathbb{E}_{M_{\mathbf{x}}}\left[d\left(\mathbf{x}, \mathbf{x}^+\right)\right]$, where $M_{\mathbf{z}}$ and $M_{\mathbf{x}}$ are measures in the embedding and input space, respectively; $\Pi(\mathcal{S}, \mathcal{S}^+)$ is the joint distribution of $\mathcal{S}$ and $\mathcal{S}^+$.
Then, the relaxed worst-case problem can be formulated as:
\begin{equation}
\theta^\ast = \min_\theta \sup_{\mathcal{S}^+\in\mathcal{D}}\mathbb{E}[\mathcal{L}_{\mathrm{task}}(\theta;\mathcal{S}^+)],
\label{eq_problem}
\end{equation}
where $\mathcal{D} = \{\mathcal{S}^+: W_c(\mathcal{S}, \mathcal{S}^+) \le \rho, W_d(\mathcal{S}, \mathcal{S}^+) \ge \eta\}$. We note that $\mathcal{D}$ 
covers a robust region that is within $\rho$ distance of $\mathcal{S}$ in the embedding space and $\eta$ distance away from $\mathcal{S}$ in the input space under the Wasserstein distance measures $W_c$ and $W_d$, respectively.

\begin{figure*}[t]
\begin{center}
\includegraphics[width=1.0\linewidth]{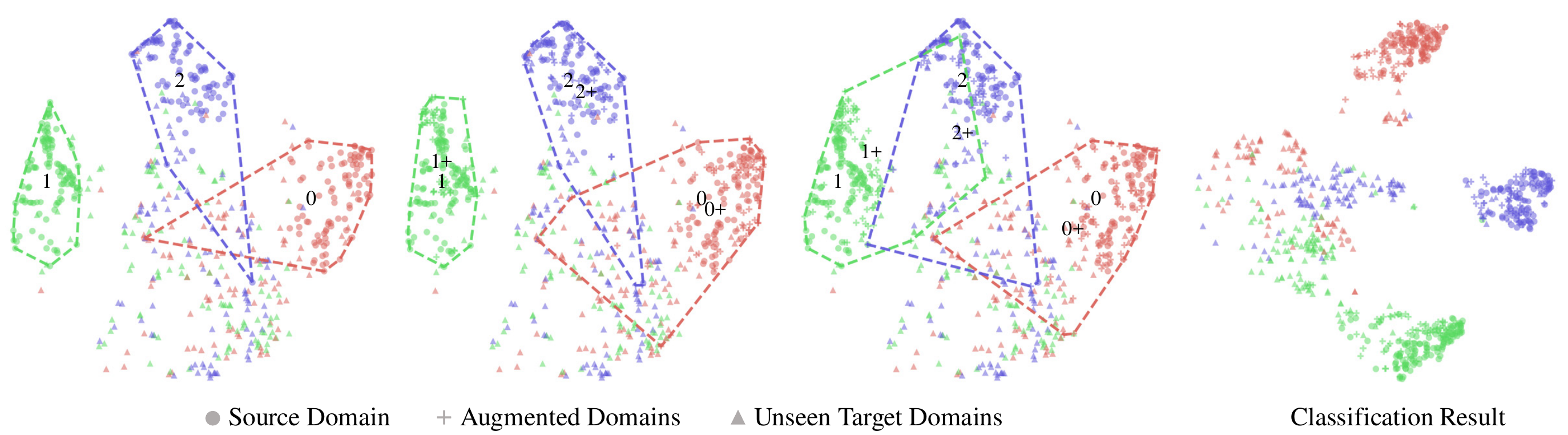}
\end{center}
\vspace{-1em}
\caption{Visualization of domains and convex hulls in the embedding space (the first three figures) and classification space (the last figure).  {\bf From left to right}: (a) source domain $\mathcal{S}$ and unseen target domains $\mathcal{T}$; (b) augmented domains $\mathcal{S}^+$ w/o $\mathcal{L}_{\mathrm{relax}}$; (c) $\mathcal{S}^+$ w/ $\mathcal{L}_{\mathrm{relax}}$; (d) the classification result of M-ADA. Different colors denote different categories. The numbers mark the corresponding cluster centers. Note that \textbf{1}: cluster center of $\mathcal{S}$; \textbf{1+}: cluster center of $\mathcal{S}^+$. Best viewed in color and zoom in for details.}
\label{fig_tsne}
\end{figure*}

For deep neural networks, Eq.~\eqref{eq_problem} is intractable with arbitrary $\rho$ and $\eta$. Consequently, we consider its Lagrangian relaxation with fixed penalty parameters $\alpha \ge 0$ and $\beta \ge 0$:
\begin{equation}\label{eq:lagrangian-duality}\nonumber
\min_{\theta}\{
\sup_{\mathcal{S}^+} \left \{ \mathbb{E}[\mathcal{L}_{\mathrm{task}}(\theta;\mathbf{x}^+) ] - W_{c,d}\right\} = \mathbb{E}[\phi_{\alpha,\beta}(\theta,\psi;\mathbf{x})]\},
\end{equation}
and we have $W_{c,d}(\mathcal{S}, \mathcal{S}^+) = \alpha W_c(\mathcal{S}, \mathcal{S}^+) -\beta W_d(\mathcal{S}, \mathcal{S}^+)$, 
$\phi_{\alpha,\beta}(\theta,\psi;\mathbf{x})= \sup_{\mathbf{x}^+}\left\{\mathcal{L}_{\mathrm{task}}(\theta;\mathbf{x}^+) -\mathcal{L}_{c,d}  \right\}$, 
and $\mathcal{L}_{c,d} = \alpha c\left(\mathbf{z}, \mathbf{z}^+\right)-\beta d\left(\mathbf{x}, \mathbf{x}^+\right)$.
Thus the problem in Eq.~\eqref{eq_problem} is transformed to minimize the robust surrogate $\phi_{\alpha,\beta}$. 

According to \cite{sinha2017certifying}, $\phi_{\alpha}$ is smooth \wrt $\theta$ if $\alpha$ is large enough and the assumption of Lipschitzian smoothness holds. Since $\psi$ and $\theta$ are independent with each other, $\phi_{\alpha,\beta}$ is still smooth \wrt $\theta$. The gradient can be computed as:
\begin{equation}\nonumber
  \nabla_\theta \phi_{\alpha,\beta}(\theta,\psi; \mathbf{x})
  = \nabla_\theta \mathcal{L}_{\mathrm{task}}(\theta; \mathbf{x}^{\star}(\mathbf{x}, \theta, \psi)),
\end{equation}
where $\mathbf{x}^\star(\mathbf{x},\theta,\psi) = \argmax_{\mathbf{x}^+} [\mathcal{L}_{\mathrm{task}}(\theta; \mathbf{x}^+)-\mathcal{L}_{c,d}] = \argmax_{\mathbf{x}^+}\mathcal{L}_\text{ADA}(\theta,\psi;\mathbf{x}^+,\mathbf{z}^+) $, which is exactly the adversarial perturbation defined in Eq.~\eqref{eq:ada}.

\section{Experiments}

%We conduct comprehensive experiments to evaluate the performance of M-ADA on single domain generalization, where models are trained on only one source domain and expected to generalize well on unseen domains. 
We begin by introducing the experimental setups and implementation details in Secs.~\ref{sec:setting} and \ref{sec:implementation}, respectively. In Sec.~\ref{sec:ablation}, we carry out detailed ablation study to validate the strength of the proposed relaxation, the efficiency of meta-learning scheme, and the selection and trade-off of key hyperparameters. In Sec.~\ref{sec:sota}, we compare M-ADA with state of the arts on benchmark datasets. %, which demonstrates its superior performance for single domain generalization. 
%Finally, 
%In Sec.~\ref{sec:few}, we further evaluate M-ADA in a popular setup of {\it few-shot domain adaptation}.%, where a few examples in the target domain are available for training.
In Sec.~\ref{sec:few}, we further evaluate M-ADA in {\it few-shot domain adaptation}.

\subsection{Datasets and Settings}\label{sec:setting}

{\bf Datasets and settings:} (1) {\it Digits} consists of five sub-datasets: MNIST \cite{lecun1998gradient}, MNIST-M \cite{ganin2015unsupervised}, SVHN \cite{netzer2011reading}, SYN \cite{ganin2015unsupervised}, and USPS \cite{denker1989advances}, and each of them can be viewed as a different domain. Each image in these datasets contains one single digit with different styles. This dataset is mainly employed for ablation studies.
We use the first 10,000 samples in the training set of MNIST for training, and evaluate models on all other domains.
(2) {\it CIFAR-10-C} \cite{hendrycks2019benchmarking} is a robustness benchmark consisting of 19 corruptions types with five levels of severities applied to the test set of CIFAR-10. The corruptions come from four main categories: noise, blur, weather and digital. Each corruption has five-level severities and ``5'' indicates the most corrupted one. All the models are trained on CIFAR-10 and evaluated on CIFAR-10-C.
(3) {\it SYTHIA} \cite{ros2016synthia} is a dataset synthesized for semantic segmentation in the context of driving scenarios. This dataset consists of the same traffic situation but under different locations (Highway, New York-like City and Old European Town are selected)  and different weather/illumination/season conditions (Dawn, Fog, Night, Spring and Winter are selected). Following the protocol in \cite{volpi2018generalizing}, we only use the images from the left front camera and 900 images are randomly sample from each source domain.

{\bf Evaluation metrics:} For Digits and CIFAR-10-C, we compute the mean 
%and standard deviation of 
accuracy on each unseen domain. For CIFAR-10-C, accuracy may not be sufficient to comprehensively evaluate the performance of models without measuring relative gain over baseline models (ERM \cite{koltchinskii2011oracle}) and relative error evaluated on the \textit{clean} dataset, \ie, the test set of CIFAR-10 without any corruption.
Inspired by the robustness metrics proposed in \cite{hendrycks2019benchmarking}, two metrics are formulated to evaluate the robustness against image corruptions in the context of domain generalization: mean Corruption Error (mCE) and Relative mCE (RmCE). They are defined as:
$\mathrm{mCE}=\frac{1}{N}\sum_{i=1}^{N} E_{i}^{f} /E_{i}^{\mathrm{ERM}}$,
$\mathrm{RmCE}=\frac{1}{N} \sum_{i=1}^{N} (E_{i}^{f}-E_{\text {clean }}^{f})/(E_{i}^{\mathrm{ERM}}-E_{\text {clean }}^{\mathrm{ERM}})$,
where $N$ is the number of corruptions. mCE is used for evaluating the robustness of the classifier $f$ compared with ERM \cite{koltchinskii2011oracle}. RmCE measures the relative robustness compared with the \textit{clean} data. For SYTHIA, we compute the standard mean Intersection Over Union (mIoU) on each unseen domain.

\begin{figure}[t]
\begin{center}
\subfigure{
\includegraphics[width=0.47\linewidth]{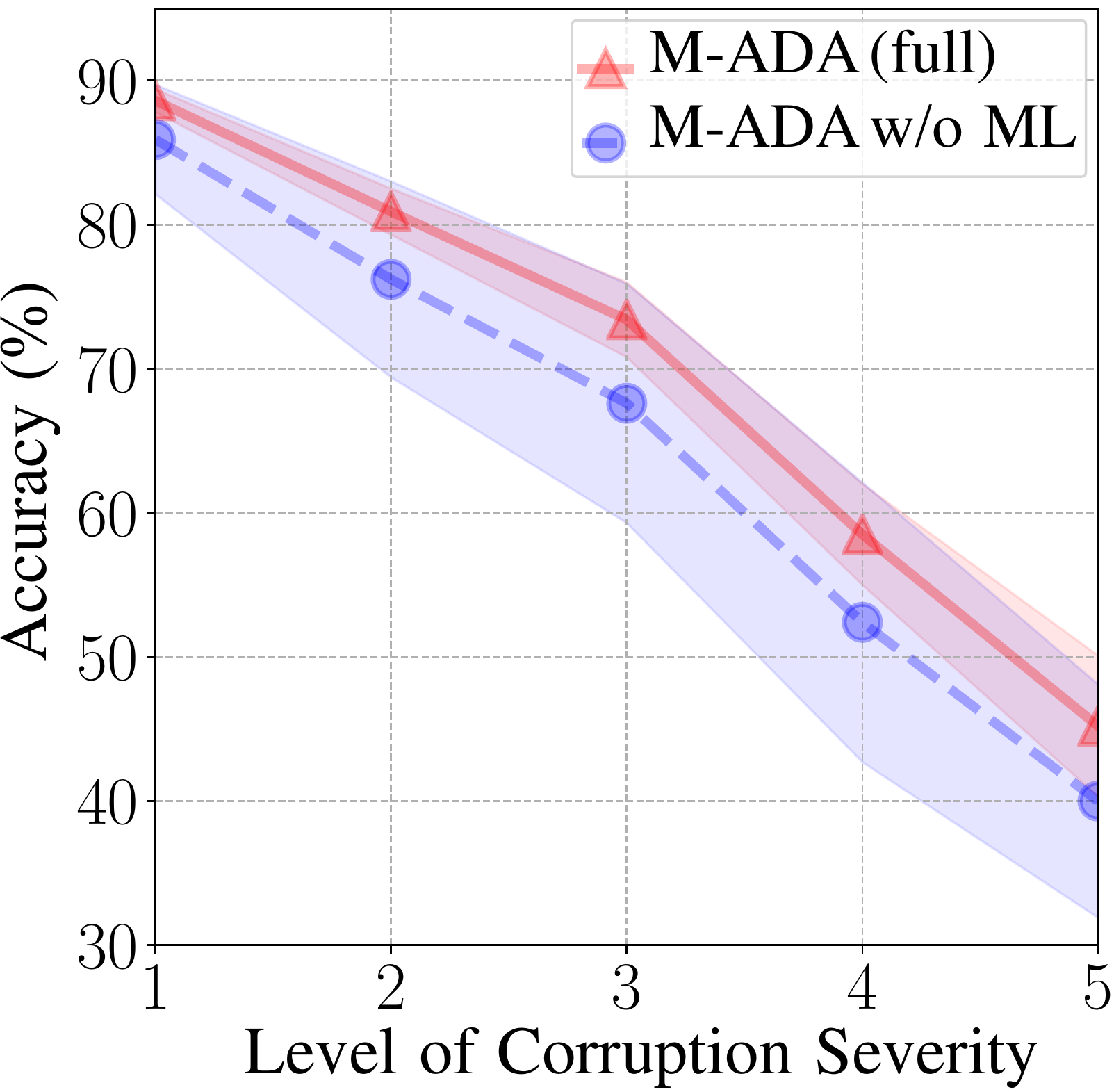}
}
\subfigure{
\includegraphics[width=0.47\linewidth]{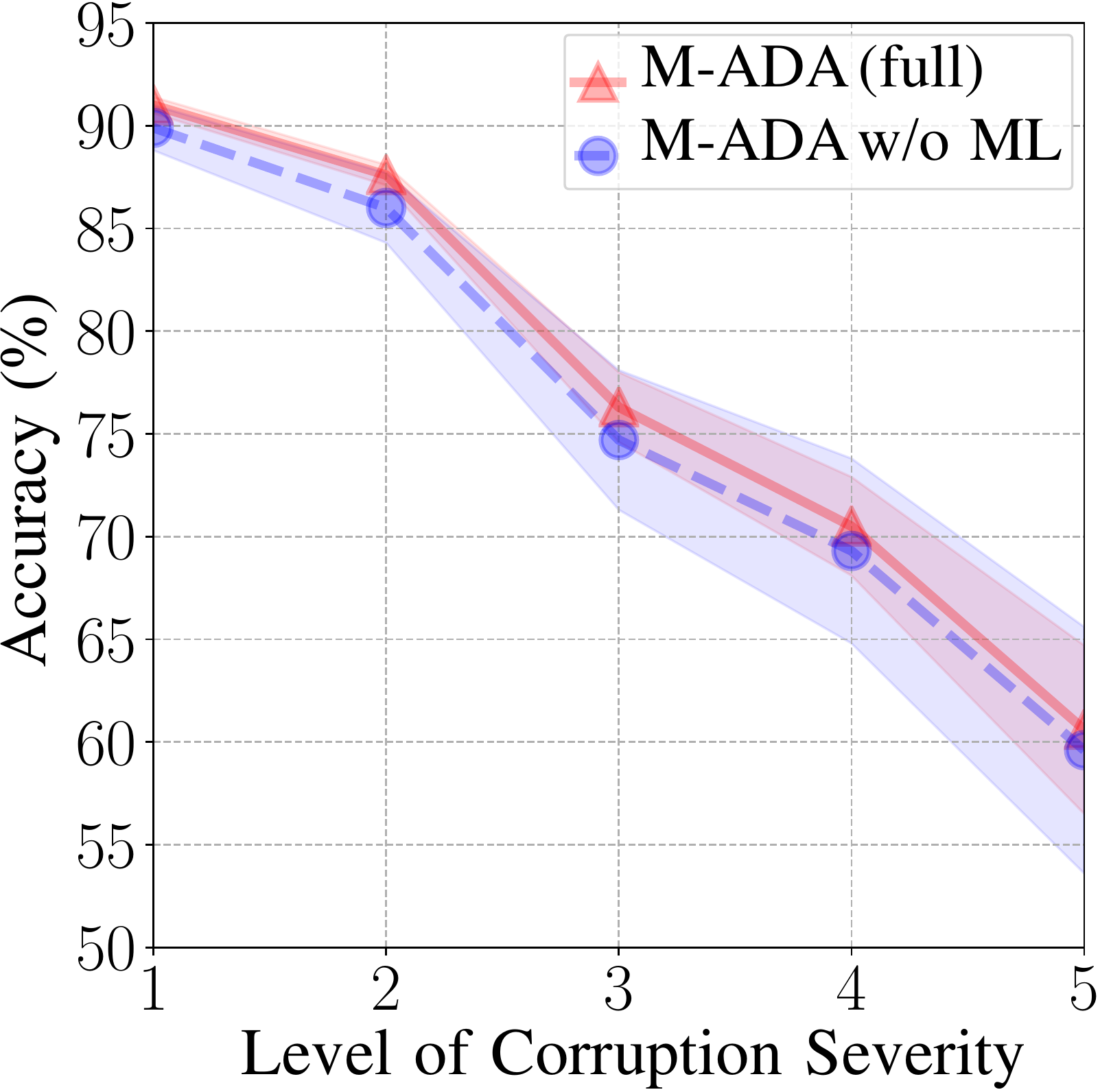}
}
\end{center}
\vspace{-1em}
\caption{Validation of meta-learning scheme. 
%on CIFAR-10-C \cite{hendrycks2019benchmarking}. 
Five levels of severity on \textit{Impulse Noise} (\textbf{left}) and \textit{Shot Noise} (\textbf{right}) are evaluated.
%ML training scheme can improve performance and stabilize the training at the same time.
}
\label{fig_cifar10meta}
\end{figure}
\begin{table}[t]
	\begin{center}
		\resizebox{.88\linewidth}{!}{
		\begin{tabular}{@{}lccccc@{}}
			\toprule
			Method&  \# of params. & Inference time  & Accuracy\\
			\hline
			GUD \cite{volpi2018generalizing} & 31.9M & 22.1ms & 55.8\% \\
			{\bf M-ADA} (full) & 4.54M &  3.07ms & 59.5\% \\
			\bottomrule
	\end{tabular}}
	\end{center}
	\caption{Efficiency comparison in single domain generalization. GUD has to learn a series of ensemble models. M-ADA leverages meta-learning scheme to achieve a single model. M-ADA outperforms GUD marginally in terms of memory, speed, and accuracy.}\label{tab:ensemble}
\end{table}

\subsection{Implementation Details}\label{sec:implementation}

{\bf Task models:} We design specific task models and employ different training strategies accordingly for the three datasets. Please refer to the supplementary material for more details. 
For {\it Digits} dataset, we use a ConvNet \cite{lecun1989backpropagation} with architecture \textit{conv-pool-conv-pool-fc-fc-softmax}. All images are resized to 32$\times$32, and the channels of MNIST and USPS are duplicated to make them as RGB images. 
We use Adam with the learning rate $\eta=0.0001$. The batch size is 32 and the total number of iterations is 10,000.
For {\it CIFAR-10-C}, we use Wide Residual Network (WRN) \cite{zagoruyko2016wide} with 16 layers and the width is 4. Following the training procedure in \cite{zagoruyko2016wide}, we use SGD with Nesterov momentum and set the batch size to 128. The initial learning rate is 0.1 with a linear decay and the number of epochs is 200. 
For {\it SYTHIA}, we use FCN-32s \cite{long2015fully} with the backbone of ResNet-50 \cite{he2016deep}. We use Adam with the learning rate $\alpha=0.0001$. We set the batch size to 8 and the number of epochs to 50.

{\bf Wasserstein Auto-Encodes:} We follow \cite{tolstikhin2018wasserstein} to implement WAEs but slightly modify architectures for dataset adaptation. The encoder and decoder are built with Fully-Connected (FC) layers for Digits dataset. We utilize two convolutional neural networks to implement the auto-encoders for CIFAR-10-C and SYTHIA. When training WAEs, we use WAE-GAN~\cite{tolstikhin2018wasserstein} to minimize the JS divergence between $P(\mathbf{e})$ and $Q(\mathbf{e}|\mathbf{x})$ in the latent space. An additional discriminator implemented by FC layers is used for distinguishing the \textit{true} points from $P(\mathbf{e})$ and \textit{fake} points from $Q(\mathbf{e}|\mathbf{x})$. Due to the space limitation, we suggest readers refer to the supplementary material for detailed setups.

\begin{figure}[t]
\begin{center}
\subfigure{
	\includegraphics[width=0.47\linewidth]{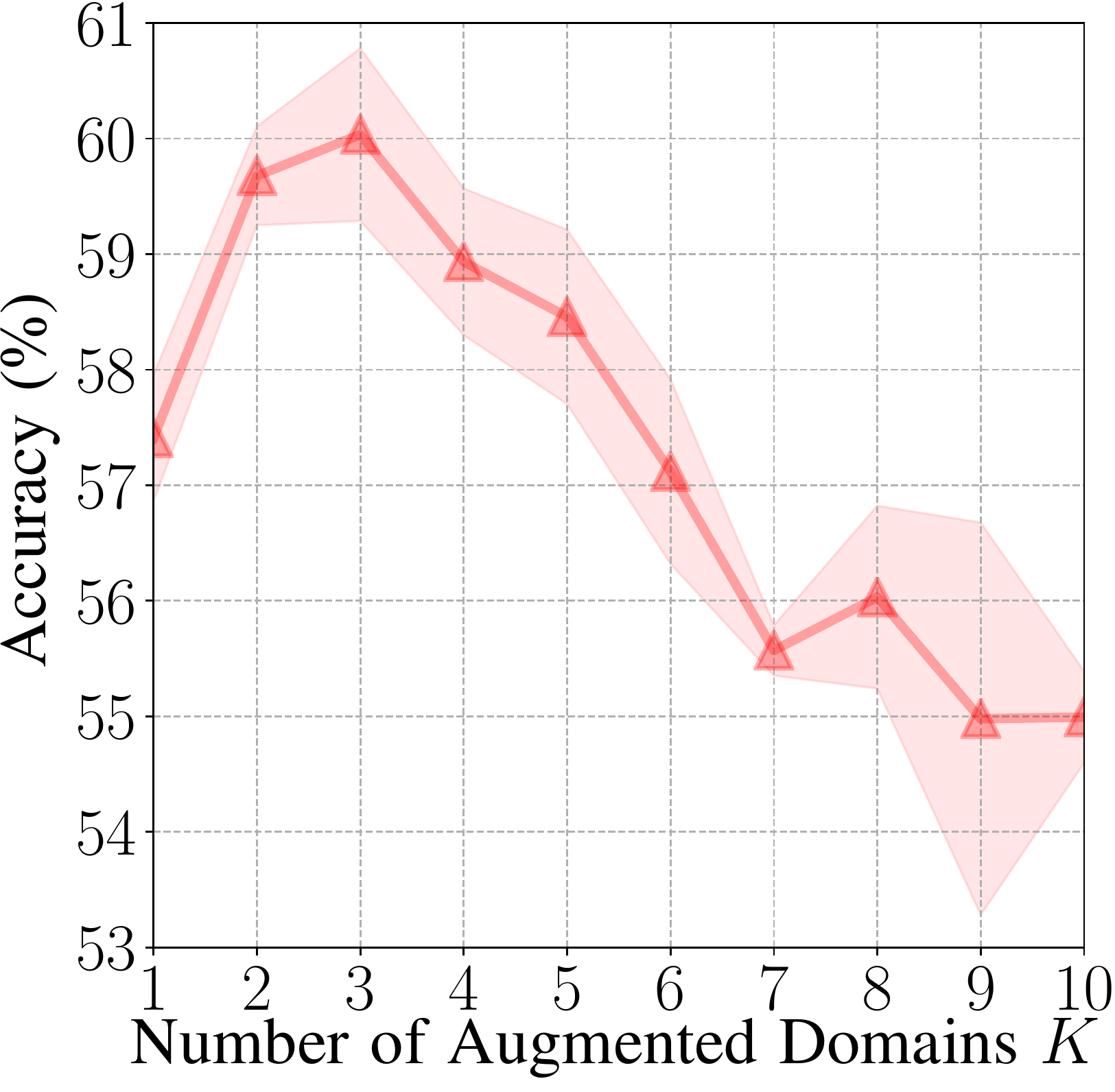}
}
\subfigure{
	\includegraphics[width=0.47\linewidth]{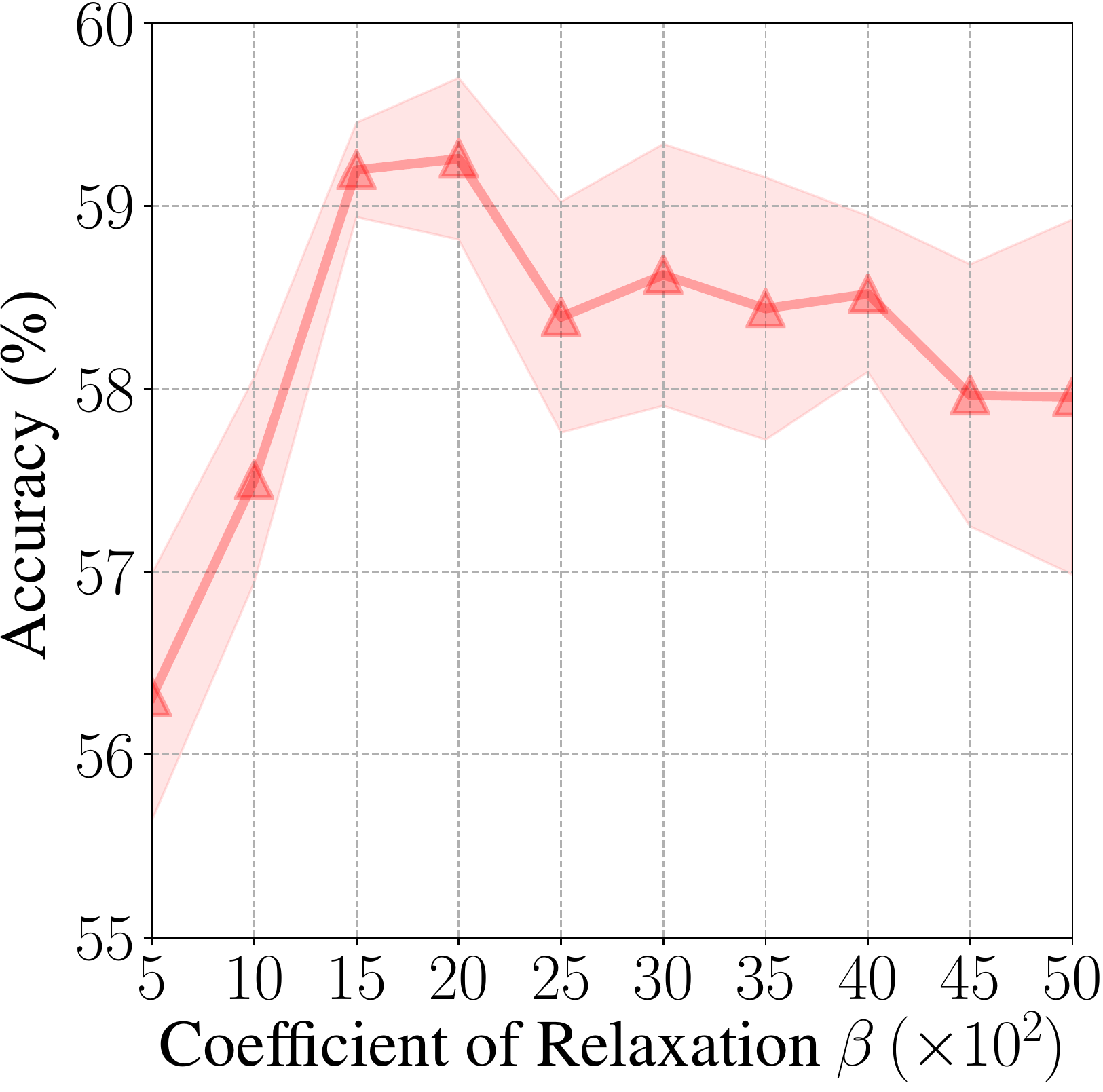}
}
\end{center}
\vspace{-1em}
\caption{Hyper-parameter tuning of $K$ and $\beta$. We set $K=3$ and $\beta=2.0\times10^3$ according to the best classification accuracy.} 
\label{fig_params}
\end{figure}
\begin{table}[t]
	\begin{center}
		\resizebox{1.\linewidth}{!}{
		\begin{tabular}{@{}lccccc@{}}
			\toprule
			Method& Level 1& Level 2& Level 3& Level 4& Level 5 \\
			\hline
			ERM \cite{koltchinskii2011oracle}& 87.8$\pm$0.1& 81.5$\pm$0.2& 75.5$\pm$0.4& 68.2$\pm$0.6& 56.1$\pm$0.8 \\
			GUD \cite{volpi2018generalizing}& 88.3$\pm$0.6& 83.5$\pm$2.0& 77.6$\pm$2.2& 70.6$\pm$2.3& 58.3$\pm$2.5 \\
			\hline
			{\bf M-ADA} (full) &\textbf{90.5$\pm$0.3}& \textbf{86.8$\pm$0.4}& \textbf{82.5$\pm$0.6}& \textbf{76.4$\pm$0.9}& \textbf{65.6$\pm$1.2} \\
			$\uparrow$ to ERM &3.08\% & 6.50\%& 9.27\%& 12.0\%& 16.9\% \\
			$\uparrow$ to GUD &2.49\%& 3.95\%& 6.31\%& 8.22\%& 12.5\% \\
			\bottomrule
	\end{tabular}}
	\end{center}
	\caption{Accuracy comparison (\%) on {\it CIFAR-10-C}. Boosts ($\uparrow$) become more significant as corruption severity level ({\bf 1-5}) increases.}\label{tab:cifar10-c_severity}
\end{table}

\subsection{Ablation Study}\label{sec:ablation}

%In this section, we conduct experiments to evaluate the effect of the proposed relaxation term $\mathcal{L}_{\mathrm{relax}}$ in Eq.~\eqref{eq:ada}, the meta-learning training scheme (Sec.~\ref{sec:3.3}) and two key parameters, $K$ and $\beta$, of M-ADA.

{\bf Validation of $\mathcal{L}_{\mathrm{relax}}$:} To give an intuitive understanding of how $\mathcal{L}_{\mathrm{relax}}$ affects the distribution of augmented domains $\mathcal{S}^+$, we use t-SNE \cite{maaten2008tsne} to visualize $\mathcal{S}^+$ with and without $\mathcal{L}_{\mathrm{relax}}$ in the embedding space. Their results are shown in Fig.~\ref{fig_tsne}~(b) and~(c), respectively. We observe that the convex hull of $\mathcal{S} \cup$ $\mathcal{S}^+$ with $\mathcal{L}_{\mathrm{relax}}$ covers an enlarged region than that of $\mathcal{S} \cup \mathcal{S}^+$ without $\mathcal{L}_{\mathrm{relax}}$. This indicates that $\mathcal{S}^+$ contains more distributional variance and better overlaps with unseen domains. Further, we compute Wasserstein distance to quantitatively measure the difference between $\mathcal{S}$ and $\mathcal{S}^+$. The distance between $\mathcal{S}$ and $\mathcal{S}^+$ with $\mathcal{L}_{\mathrm{relax}}$ is 0.078, while if $\mathcal{L}_{\mathrm{relax}}$ is not employed, the distance decreases to 0.032, indicating an improvement of $58.9\%$ by introducing $\mathcal{L}_{\mathrm{relax}}$. These results demonstrate that $\mathcal{L}_{\mathrm{relax}}$ is capable of pushing $\mathcal{S}^+$ away from $\mathcal{S}$, which guarantees significant domain transportation in the input space.

{\bf Validation of meta-learning scheme:} The comparisons of M-ADA with and without meta-learning (ML) scheme are presented in Tabs.~\ref{tab_digits} and~\ref{tab:cifar10_s5}. We observe that with the help of this meta-learning scheme, the results on average accuracy of Digits and CIFAR-10-C are improved by 0.94\% and 1.37\%, respectively. Specially, the results of two kinds of unseen corruptions are shown in Fig.~\ref{fig_cifar10meta}. As seen, M-ADA can significantly reduce variance and yield better performance across all levels of severity. The experimental results prove that the meta-learning scheme plays a key role to improve the training stability and classification accuracy. This is extremely important when performing adversarial domain augmentation in challenging conditions.

{\bf Hyper-parameter tuning of $K$ and $\beta$:} We study the effect of two important hyper-parameters of M-ADA: the number of augmented domains ($K$) and the deviation between the source and augmented domain ($\beta$). We plot the accuracy curve under different $K$ and $\beta$ in Fig.~\ref{fig_params}. In Fig.~\ref{fig_params}~(left), we find that the accuracy reaches the summit when $K=3$ and keeps falling with $K$ increasing. This is due to the fact that excessive adversarial samples above a certain threshold will increase the instability and degrade the robustness of the model. 
% {\color{blue} Since the distance between the augmented and source domain increases as $K$ increases, a large $K$ may break down the constraint of semantic consistency yielding inferior model training.}
In Fig.~\ref{fig_params}~(right), we observe that the accuracy reaches the summit when $\beta=2.0 \times 10^3$ and drops slightly when $\beta$ increases. This is because large $\beta$ will produce domains too far way from the source $\mathcal{S}$ and even reach out of the manifold in the embedding space.

\subsection{Evaluation of Single Domain Generalization}\label{sec:sota}

\begin{table}[t]
\begin{center}
\resizebox{1.\linewidth}{!}{
\begin{tabular}{@{}lccccc@{}}
\toprule
%Method               & SVHN~\cite{netzer2011reading}  & MNIST-M~\cite{ganin2015unsupervised} & SYN~\cite{ganin2015unsupervised} & USPS~\cite{denker1989advances} & Avg.  \\
Method               & SVHN  & MNIST-M & SYN & USPS & Avg.  \\
\hline
ERM \cite{koltchinskii2011oracle}                  &  27.83       & 52.72 & 39.65 & 76.94 & 49.29\\
CCSA \cite{motiian2017unified} & 25.89 & 49.29 & 37.31  & 83.72   & 49.05   \\
d-SNE \cite{xu2019dsne} &26.22      &  50.98  & 37.83    &\textbf{93.16 }& 52.05 \\
JiGen  \cite{carlucci2019jigasaw} & 33.80     &  57.80   &43.79  & 77.15   & 53.14 \\
GUD \cite{volpi2018generalizing} & 35.51   &  60.41 & 45.32 & 77.26 &  54.62\\
\hline
M-ADA w/o $\mathcal{L}_{\mathrm{relax}}$ & 37.33 & 61.43   & 45.58 & 77.37 & 55.43   \\
M-ADA w/o $\mathcal{L}_{\mathrm{const}}$ & 41.36  & 67.28  & 47.94 & 78.22 &  58.70 \\
M-ADA w/o ML & 41.45  & 67.86 & 48.76 & 76.12 &  58.55\\
{\bf M-ADA} (full) &\textbf{42.55} & \textbf{67.94} & \textbf{48.95} &78.53 & \textbf{59.49}\\
\bottomrule
\end{tabular}}
\end{center}
\caption{Single domain generalization comparison (\%) on {\it Digits}. Models are trained on MNIST. The variant (w/o $\mathcal{L}_{\mathrm{relax}}$) has the most significant performance decrease, indicating it is crucial to perform Wasserstein relaxation for single domain generalization.}\label{tab_digits}
\end{table}

We compare our method with the following five state-of-the-art methods. 
(1) Empirical Risk Minimization (ERM) \cite{vapnik1998statistical,koltchinskii2011oracle} are models trained with cross-entropy loss, without any auxiliary loss and data augmentation scheme.
(2) CCSA~\cite{motiian2017unified} uses semantic alignment to regularize the learned feature subspace for domain generalization. 
(3) d-SNE~\cite{xu2019dsne} minimizes the largest distance between the samples from the same class and maximizes the smallest distance between the samples from different classes.
(4) GUD~\cite{volpi2018generalizing} proposes an adversarial data augmentation method for single domain generalization, which is the related work to M-ADA.
(5) JiGen~\cite{carlucci2019jigasaw} learns to classify and predict the order of shuffled image patches at the same time for domain generalization.

{\bf Comparison on Digits:} We train all models on MNIST and test them on unseen domains, \ie, MNIST-M, SVHN, SYN, and USPS. We report the results in Tab.~\ref{tab_digits}. We observe that M-ADA outperforms GUD with a large margin on SVHN, MNIST-M and SYN. The improvement on USPS is not as significant as those on other domains, mainly due to its great similarity with MNIST. On the contrary, CCSA and d-SNE obtain large improvements on USPS but perform poorly on other ones. We also compare M-ADA with an ensemble model of GUD, which aggregates prediction results of several models under different semantic constraints. Results are shown in Tab.~\ref{tab:ensemble}. As seen, M-ADA outperforms GUD ensemble models in terms of generalization accuracy but with much less model parameters and even faster inference speed. The strong results, once again, testify the efficiency of the proposed learning to learn framework.

\begin{figure}[t]
\begin{center}
\includegraphics[width=1.0\linewidth]{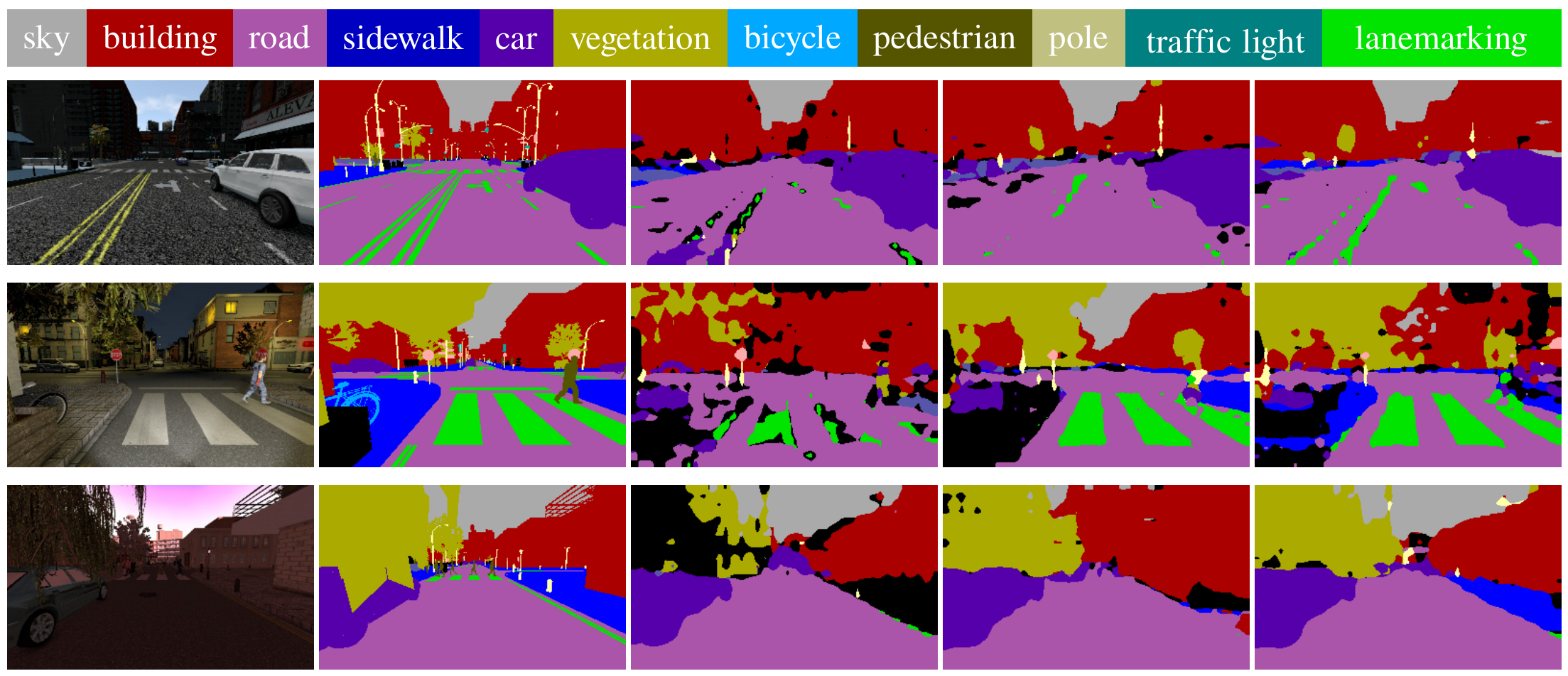}
\end{center}
\caption{Examples of semantic segmentation on {\it SYNTHIA}~\cite{ros2016synthia}. {\bf From left to right:} (a) images from unseen domains; (b) ground truth; (c) results of ERM~\cite{koltchinskii2011oracle}; (d) results of GUD~\cite{volpi2018generalizing}; and (e) results of M-ADA. Best viewed in color and zoom in for details.}
\label{fig:segmentation}
\end{figure}

{\bf Comparison on CIFAR-10-C:} We train all models on the clean data, {\it i.e.}, CIFAR-10, and test them on the corruption data, {\it i.e., CIFAR-10-C}. In this case, there are totally $19$ unseen testing domains. Results on CIFAR-10-C across five levels of corruption severity are shown in Tab.~\ref{tab:cifar10-c_severity}. As seen, The gap between GUD and M-ADA gets larger with the level of severity increasing, and M-ADA can significantly reduce standard deviations across all levels. In addition, we present the result of each corruption with the highest severity in Tab.~\ref{tab:cifar10_s5}. We observe that M-ADA substantially outperforms other methods on most corruptions. Specially, in several corruptions such as \textit{Snow}, \textit{Glass blur}, \textit{Pixelate} and corruptions related with \textit{Noise}, M-ADA outperforms ERM \cite{koltchinskii2011oracle} with more than 10\%. More importantly, M-ADA has the lowest values on mCE and RmCE, indicating its strong robustness against image corruptions. 

{\bf Comparison on SYTHIA:} In this experiment, Highway is the source domain, and New York-like City together with Old European Town are unseen target domains. We report semantic segmentation results in Tab.~\ref{tab:seg} and show some examples in Fig.~\ref{fig:segmentation}. Unseen domains are from different locations and other conditions. We observe that M-ADA obtains the highest values on average mIoUs across three source domains, suggesting its capability of coping with changes of locations, weather and time. Improvements over ERM \cite{koltchinskii2011oracle} and GUD \cite{volpi2018generalizing} are not significant compared with the other two datasets, mainly owing to the limited number of training images and high reliance of unseen domains.

\begin{table*}[t]
\begin{center}
\resizebox{1.\linewidth}{!}{
\begin{tabular}{@{}lccccccccccccccc@{}}
\toprule
&\multicolumn{3}{c}{Weather}& \multicolumn{3}{c}{Blur}& \multicolumn{3}{c}{Noise}&\multicolumn{3}{c}{Digital}& \\
\cmidrule(lr){2-4} \cmidrule(lr){5-7} \cmidrule(lr){8-10} \cmidrule(lr){11-13}
&Fog& Snow& Frost& Zoom& Defocus& Glass& Speckle& Shot& Impulse& Jpeg& Pixelate& Spatter& Avg.& mCE &RmCE \\
\hline
ERM \cite{koltchinskii2011oracle}&65.92& 74.36& 61.57& 59.97& 53.71& 49.44& 41.31& 35.41& 25.65& 69.90& 41.07& 75.36& 56.15&1.00&1.00  \\
CCSA \cite{motiian2017unified}&66.94& 74.55& 61.49& 61.96& 56.11& 48.46& 40.12& 33.79& 24.56& 69.68& 40.94& 77.91& 56.31& 0.99&0.99 \\
d-SNE \cite{xu2019dsne}&65.99& 75.46& 62.25& 58.47& 53.71& 50.48& 45.30& 39.93& 27.95& 70.20& 38.46& 73.40& 56.96&0.99 &1.00 \\
% JiGen  \cite{carlucci2019jigasaw}&& & & & & & & & & & & & & & \\
GUD \cite{volpi2018generalizing}&68.29& 76.75& 69.94& 62.95& 56.41& 53.45& 38.45& 36.87& 22.26& 74.22& \textbf{53.34}& 80.27& 58.26&0.97&0.95 \\
\hline
M-ADA w/o $\mathcal{L}_{\mathrm{relax}}$ &66.99& 80.09& 74.93& 54.15& 44.67& 60.57& 59.88& 59.18& 43.46& 76.45& 53.13& 80.75& 61.92&0.90&0.86 \\
M-ADA w/o ML&67.68& \textbf{80.91}& 76.20& 65.70& 56.87&\textbf{ 62.14}& 60.01& 59.63& 40.04& \textbf{77.62}& 52.49& \textbf{81.02}& 64.22&0.85 &0.80  \\
{\bf M-ADA} (full)&\textbf{69.36}& 80.59& \textbf{76.66}& \textbf{68.04}& \textbf{61.18}& 61.59& \textbf{60.88}& \textbf{60.58}& \textbf{45.18}& 77.14& 52.25& 80.62& \textbf{65.59}&\textbf{0.82}&\textbf{0.77}\\
\bottomrule
\end{tabular}}
\end{center}
\caption{Robustness comparison on {\it CIFAR-10-C} \cite{hendrycks2019benchmarking}. The models are generalized from the clean data to different corruptions. We report the classification accuracy (\%) of 19 corruptions (only 12 are shown) under the corruption level of ``5'' (the severest). We also report the mean Corruption Error (mCE) and relative mCE (RmCE) in the last two columns. The lower the better for mCE and RmCE.  
%For each method, we also report mCE and RmCE (Sec.~\ref{sec:setting}) computed over 19 corruptions in the \textbf{last two columns}. The lower the better for mCE and RmCE.
}
\label{tab:cifar10_s5}
\end{table*}
\begin{table*}[th]
\begin{center}
\resizebox{.86\linewidth}{!}{
\begin{tabular}{@{}llccccccccccc@{}}
\toprule
& &\multicolumn{5}{c}{New York-like City}& \multicolumn{5}{c}{Old European Town}& \\
\cmidrule(lr){3-7} \cmidrule(lr){8-12}
Source Domain& Method &Dawn&Fog&Night&Spring&Winter& Dawn&Fog&Night&Spring&Winter& Avg. \\
\hline
\multirow{3}*{Highway/Dawn}&ERM \cite{koltchinskii2011oracle}& 27.80&2.73&0.93&6.80&1.65&52.78&31.37&15.86&33.78&13.35&18.70 \\
&GUD \cite{volpi2018generalizing}& 27.14&4.05&1.63&7.22&2.83&52.80&34.43&18.19&33.58&14.68&19.66 \\
&{\bf M-ADA}& \textbf{29.10}&\textbf{4.43}&\textbf{4.75}&\textbf{14.13}&\textbf{4.97}&\textbf{54.28}&\textbf{36.04}&\textbf{23.19}&\textbf{37.53}&\textbf{14.87}&\textbf{22.33} \\
\hline
\multirow{3}*{Highway/Fog}&ERM \cite{koltchinskii2011oracle}& 17.24&34.80&12.36&26.38&11.81&33.73&55.03&26.19&41.74&12.32&27.16 \\
&GUD \cite{volpi2018generalizing}& 18.75&\textbf{35.58}&\textbf{12.77}&26.02&13.05&37.27&\textbf{56.69}&28.06&\textbf{43.57}&\textbf{13.59}&28.53 \\
&{\bf M-ADA}& \textbf{21.74}&32.00&9.74&\textbf{26.40}&\textbf{13.28}&\textbf{42.79}&56.60&\textbf{31.79}&42.77&12.85&\textbf{29.00} \\
\hline
\multirow{3}*{Highway/Spring}&ERM \cite{koltchinskii2011oracle}& 26.75&26.41&18.22&32.89&24.60&51.72&51.85&35.65&54.00&28.13&35.02 \\
&GUD \cite{volpi2018generalizing}& 28.84&29.67&20.85&35.32&27.87&52.21&\textbf{52.87}&35.99&55.30&\textbf{29.58}&36.85 \\
&{\bf M-ADA}& \textbf{29.70}&\textbf{31.03}&\textbf{22.22}&\textbf{38.19}&\textbf{28.29}&\textbf{53.57}&51.83&\textbf{38.98}&\textbf{55.63}&25.29&\textbf{37.47} \\
\bottomrule
\end{tabular}}
\end{center}
\caption{Semantic segmentation comparison on \textit{SYNTHIA} \cite{ros2016synthia}. The models are generalized from one source domain to many unseen environment settings. We report the standard mean Intersection Over Unions (mIoUs) and demonstrate visual results in Fig.~\ref{fig:segmentation}.}\label{tab:seg}
\end{table*}
\begin{table}[th]
	\begin{center}
		\resizebox{.94\linewidth}{!}{
		\begin{tabular}{@{}lccccc@{}}
			\toprule
			Method & $\lvert  \mathcal{T} \rvert$  & U \ $\rightarrow$ M & M \ $\rightarrow$ S &  S \ $\rightarrow$ M & Avg.\\
			\hline
			I2I \cite{murez2018image} & \multirow{5}{*}{All}     & 92.20 & - & 92.10 & -\\
			DIRT-T \cite{shu2018dirt} & &   - &54.50 & \textbf{99.40}&-\\
			SE \cite{french2017self}& &    \textbf{98.07} & 13.96 &99.18& 70.40\\
			SBADA \cite{russo2018source} & &   97.60 & \textbf{61.08} &76.14& 78.27\\
			G2A \cite{sankaranarayanan2018generate}&    & 90.80 & - & 92.40& -\\
			\hline
			FADA \cite{motiian2017few} & 7  & 91.50& 47.00 & 87.20&  75.23\\
			CCSA \cite{motiian2017unified} & 10  & 95.71 &37.63 & 94.57 & 75.97\\
			\hline
			\multirow{3}{*}{\bf M-ADA} & 0  & 71.19 & 36.61 & 60.14 & 55.98\\
			&7 & 92.33 & 56.33  & 89.90  & 79.52\\
			&10 &  93.67  & 57.16 & 91.81 & \textbf{80.88}\\
			\bottomrule
		\end{tabular}}
	\end{center}
\caption{Few-shot domain adaptation comparison on {\it MNIST(M), USPS(U), and SVHN(S)} in terms of accuracy (\%).  $\lvert \mathcal{T} \rvert$ denotes the number of target samples (per class) used during model training.}\label{tab:few-shot}
\end{table}

\subsection{Evaluation of Few-Shot Domain Adaptation}\label{sec:few}

{\bf Settings:} Although M-ADA is designed for single domain generalization, as mentioned in Sec.~\ref{sec:3.3}, we also show that M-ADA can be easily applied for few-shot domain adaptation~\cite{motiian2017few}.
In few-shot learning, models are usually first pre-trained on the source domain $\mathcal{S}$ and then fine-tuned on the target domain $\mathcal{T}$. More specifically, we first train M-ADA on $\mathcal{S}$ using all training images. Then we randomly pick out 7 or 10 images per class from $\mathcal{T}$. These images are used to fine-tune the pre-trained model with a learning rate of 0.0001 and a batch size of 16.

{\bf Discussions:} We compare our method with the state-of-the-art methods for few-shot domain adaptation. We also report the results of some unsupervised methods which use images in the target domain for training. Results on MNIST, USPS, and SVHN are shown in Tab.~\ref{tab:few-shot}. We observe that M-ADA obtains competitive results compared with FADA~\cite{motiian2017few} and CCSA~\cite{motiian2017unified}. And M-ADA also outperforms several unsupervised methods which take advantage of unlabeled images from the target domain.
More importantly, we note that both FADA~\cite{motiian2017few} and CCSA~\cite{motiian2017unified} are trained in a manner where samples from $\mathcal{S}$ and $\mathcal{T}$ are strongly coupled. This means that when the target domain changes, an entirely new model has to be trained. On the other hand, for a new target domain, M-ADA only needs to fine-tune the pre-trained model with a few samples within a small number of iterations. This demonstrates the high flexibility of M-ADA.

%In Digits, we evaluate the ability of generalizing to unseen datasets. In CIFAR-10-C, we evaluate the robustness on different types of image corruptions. In SYTHIA, we evaluate whether our method is robust to image segmentation with the change of locations, weather and time.
%Experiments of the same dataset are shared with the same model architecture and training parameters. ERM and Baseline are models trained only with cross-entropy loss.

%\textbf{Datasets.} We evaluate the proposed approach on three standard datasets: Digits, CIFAR-10-C and SYTHIA, which are widely used in domain adaptation.
\section{Conclusion}
In this paper, we present Meta-Learning based Adversarial Domain Augmentation (M-ADA) to address the problem of single domain generalization. 
The core idea is to use a meta-learning based scheme for efficiently organizing the training of augmented ``fictitious'' domains, which are OOD from source domain and created by adversarial training. In the future, we expect to further extend our work to address regression problems, or knowledge transferring in multimodal learning.

{\small
\bibliographystyle{ieee_fullname}
\bibliography{egbib,egbib_peng}
}

\clearpage
\begin{appendices}
%%%%%%%%% BODY TEXT

% \section{Source Code Instructions}
% We provide the source code in the folder of  ``\textbf{code}''. To reproduce our results on Digits dataset, please run:

% \centerline{\textbf{sh main.sh}} 

% Details of other files and folders are as follow:
% \begin{itemize}
% 	\item \textbf{data:} MNIST~\cite{lecun1998gradient} and test sets of SVHN~\cite{netzer2011reading}, MNIST-M~\cite{ganin2015unsupervised}, SYN~\cite{ganin2015unsupervised}, and USPS~\cite{denker1989advances}.
% 	\item \textbf{models:} definition of the classification model, WAE, and pre-trained models.
% 	\item \textbf{utils:} loss functions and other supportive functions.
% 	\item \textbf{train.py:} the source code to train our model.
% 	\item \textbf{test.py:} the source code to deploy the trained model.
% \end{itemize}

% The code is built on Pytorch 1.1.0 with CUDA 9.0. MetaNN 0.1.5 is used to implement meta-learning and Scipy 1.2.1 is used to process images. We use one NVIDIA Titan XP GPU on CentOS 6.10.
% \section{Supplementary Material}

\section{Experimental Details}

{\bf Task models:} We design specific task models and employ different training strategies for the three datasets according to their characteristics.

In Digits dataset, the model architecture is \textit{conv-pool-conv-pool-fc-fc-softmax}.
There are two 5 $\times$ 5 convolutional layers with 64 and 128 channels respectively. Each convolutional layer is followed by a max pooling layer with the size of 2 $\times$ 2. The size of the two Fully-Connected (FC) layers is 1024 and the size of the softmax layer is 10.

In CIFAR-10-C~\cite{hendrycks2019benchmarking}, we use Wide Residual Network (WRN)~\cite{zagoruyko2016wide} with 16 layers and the width is 4. The first
layer is a 3$\times$3 convolutional layer. It converts the original image with 3 channels to feature maps of 16 channels. Then the
features go through three groups of 3$\times$3 convolutional layers. Each group consists of two blocks and each block is composed of two convolutional layers with the same number of channels. And their channels are \{64, 128, 256\} respectively. Each convolutional layer is followed by batch normalization (BN). 
An average pooling layer with the size of 8 $\times$ 8 is appended to the output of the third group. Finally, a softmax layer with the size of 10 predicts the distribution over classes.

In SYTHIA~\cite{ros2016synthia}, we use FCN-32s~\cite{long2015fully} with a backbone of ResNet-50~\cite{he2016deep}. The model begins with ResNet-50. 
%where the final classifier layer is discarded, and all FC layers are converted to convolutional layers. 
1$\times$1 convolutional layer with 14 channels is appended to predict scores for each class at each of the coarse output locations. A deconvolution layer is followed to up-sample the coarse outputs to the original size through bilinear interpolation.

{\bf Wasserstein Auto-Encodes:} We follow~\cite{tolstikhin2018wasserstein} to implement WAEs but slightly modifying architectures for the three datasets according to their characteristics.

\begin{figure*}[t]
\begin{center}
\subfigure{
	\includegraphics[width=0.20\linewidth]{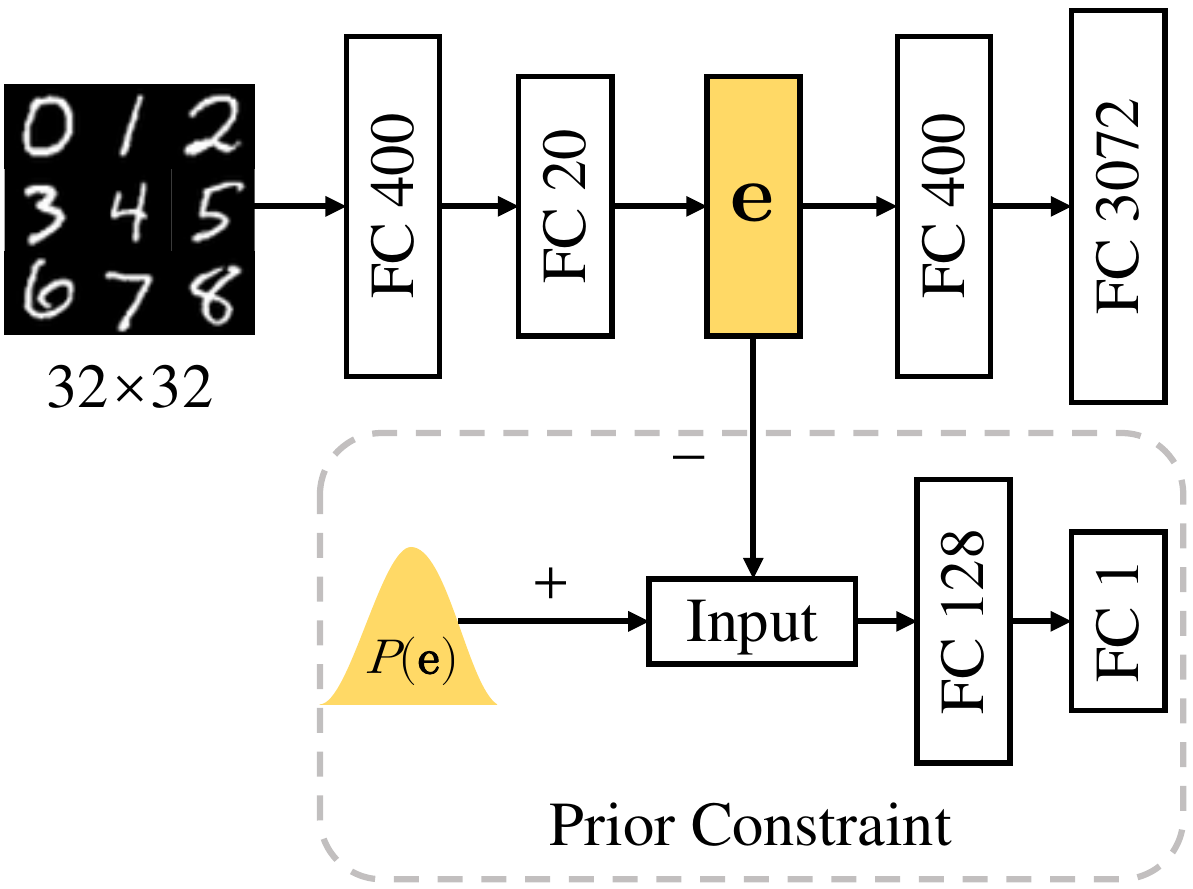}
}
\subfigure{
	\includegraphics[width=0.40\linewidth]{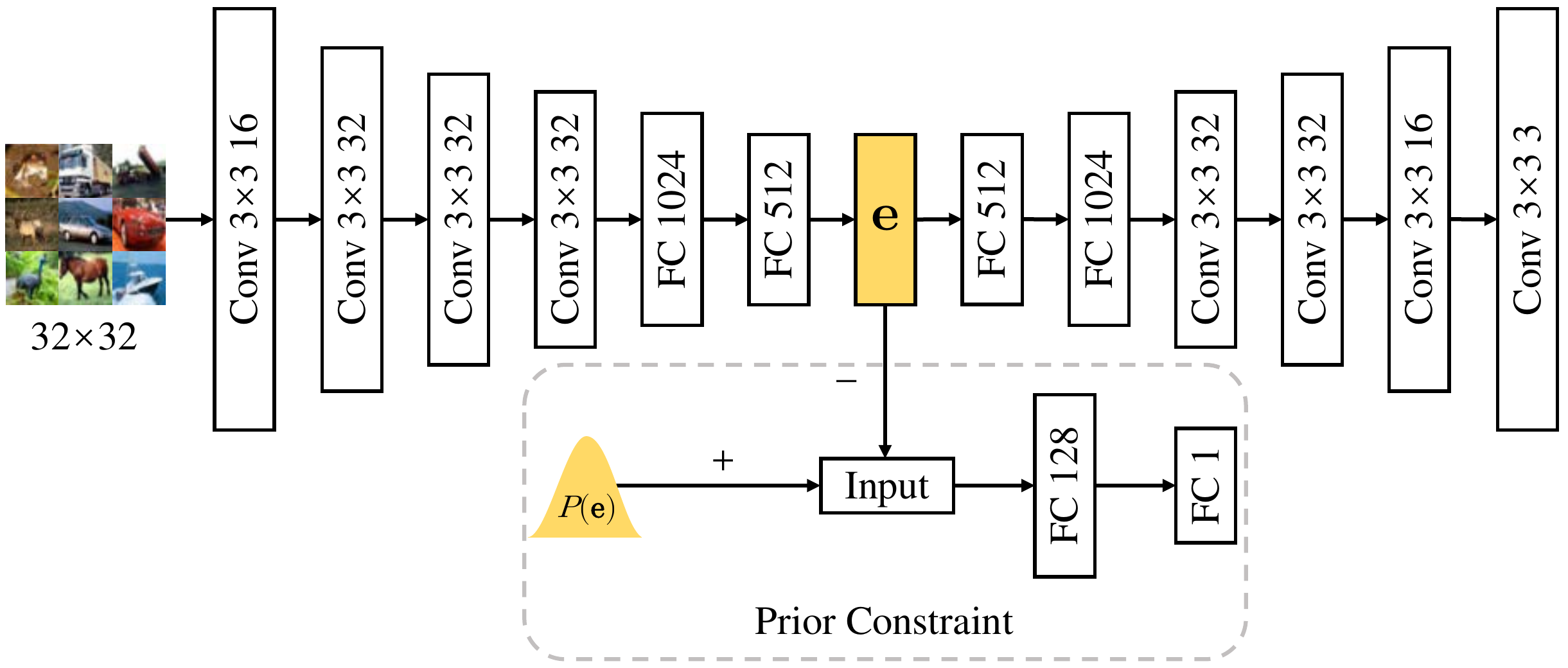}
}
\subfigure{
	\includegraphics[width=0.36\linewidth]{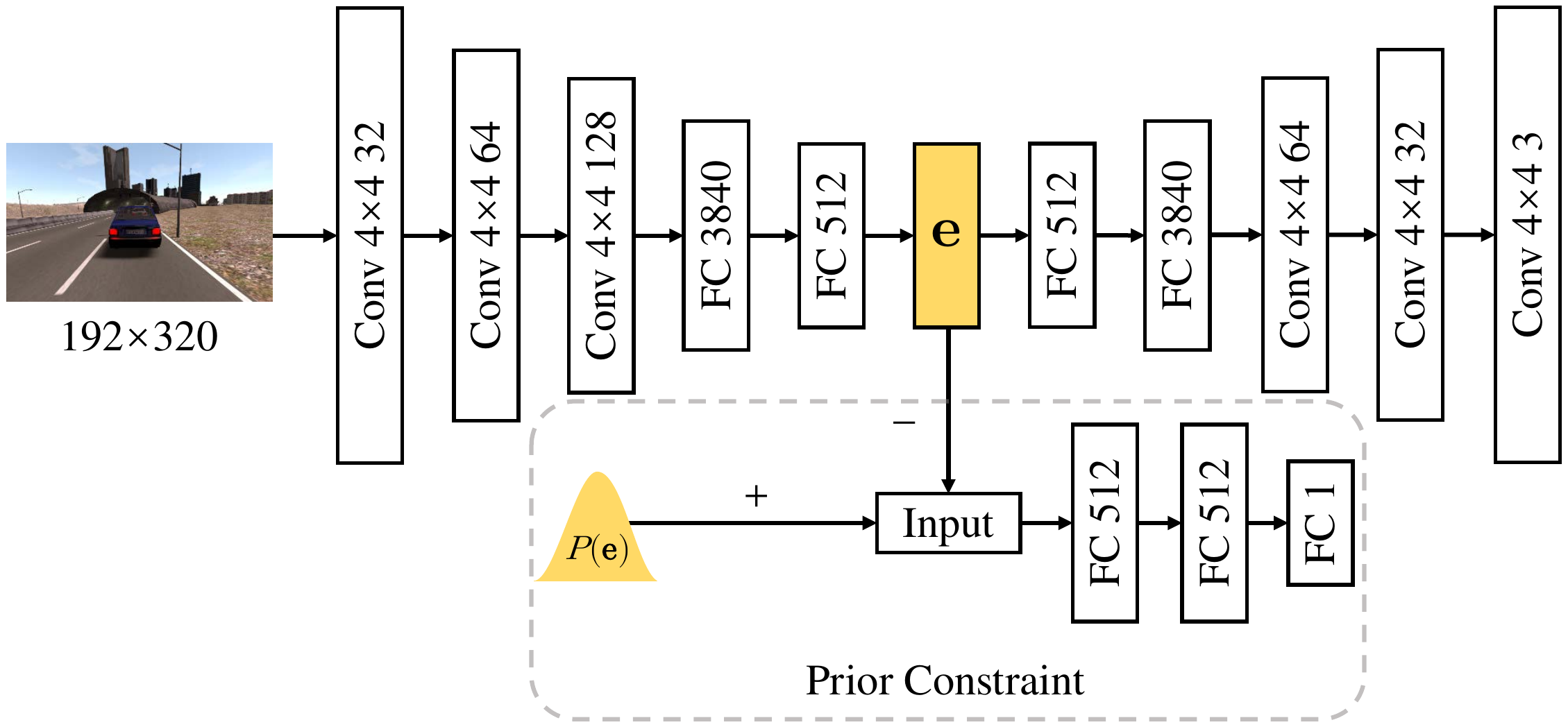}
}
\end{center}
\vspace{-1em}
\caption{Architectures of WAEs. \textbf{From left to right:} (a) WAE for {\it Digits }; (b) WAE for {\it CIFAR-10-C}~\cite{hendrycks2019benchmarking}; and (c) WAE for {\it SYTHIA}~\cite{ros2016synthia}.
Note that ``\textbf{+}'': positive samples for discriminator; ``\textbf{-}'': negative samples for discriminator. } 
\label{fig_waes_supp}
\end{figure*}

% {\it Digits }，{\it CIFAR-10-C}~\cite{hendrycks2019benchmarking}，{\it SYTHIA}~\cite{ros2016synthia}

In Digits dataset, the encoder and decoder are built with FC layers. The encoder consists of two FC layers with the size of 400 and 20 respectively. Accordingly, the decoder consists of two FC layers with the size of 400 and 3072 respectively.
The discriminator consists of two FC layers with the size of 128 and 1 respectively. 
The architecture of is shown in Fig.~\ref{fig_waes_supp}~(a).

In CIFAR-10-C~\cite{hendrycks2019benchmarking}, the encoder begins with four convolutional layers with the channels of \{16, 32, 32, 32\}. And two FC layers with the size of 1024 and 512 are followed. Accordingly, the decoder begins with two FC layers with the size of 512 and 1024 respectively. And four deconvolution layers with the channels of \{32, 32, 16, 3\} are followed.
Each layer is followed by BN except for the final layer of the decoder.
The discriminator consists of two FC layers with the size of 128 and 1 respectively. The architecture is shown in Fig.~\ref{fig_waes_supp}~(b).

In SYTHIA~\cite{ros2016synthia}, the encoder begins with three convolutional layers with the channels of \{32, 64, 128\}. And two FC layers with the size of \{3840, 512\} are followed. Accordingly, the decoder begins with two FC layers with the size of \{512, 3840\}. And three deconvolution layers with the channels of \{64, 32, 3\} are followed. Each layer is followed by BN except for the final layer of the decoder. 
The discriminator consists of three FC layers with the size of \{512, 512, 1\}. The architecture is shown in Fig.~\ref{fig_waes_supp}~(c).

We apply the Adam optimizer in training WAEs. The learning rate is 0.001 for Digits and 0.0001 for both CIFAR-10-C and SYTHIA.
The training epoches is 20 for Digits, 100 for CIFAR-10-C~\cite{hendrycks2019benchmarking}, and 200 for SYTHIA~\cite{ros2016synthia}.

\section{Additional Experimental Results}

\subsection{Ablation Study}

\begin{figure*}[t]
\begin{center}
\subfigure{
\includegraphics[width=0.23\linewidth]{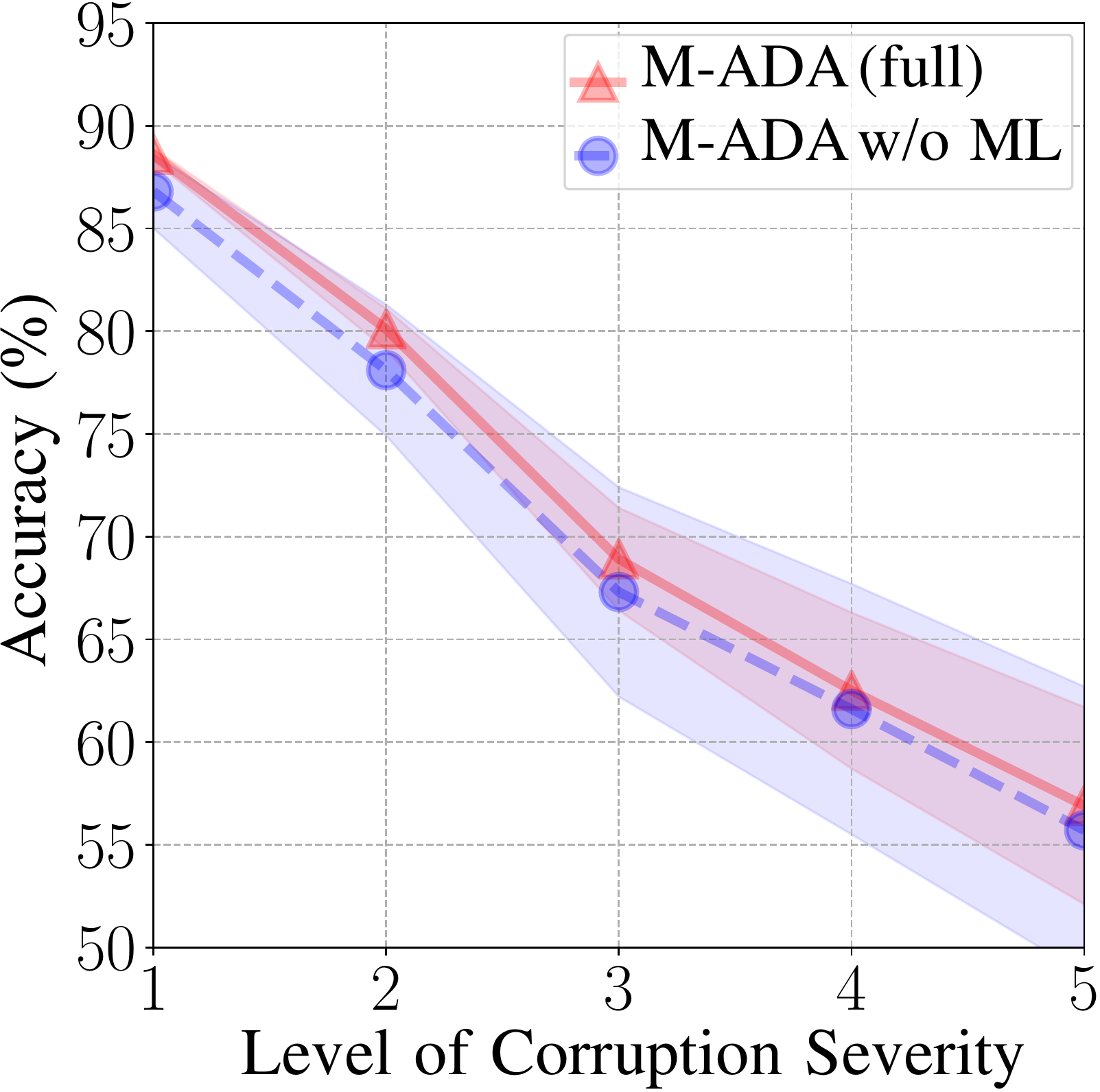}
}
\subfigure{
\includegraphics[width=0.23\linewidth]{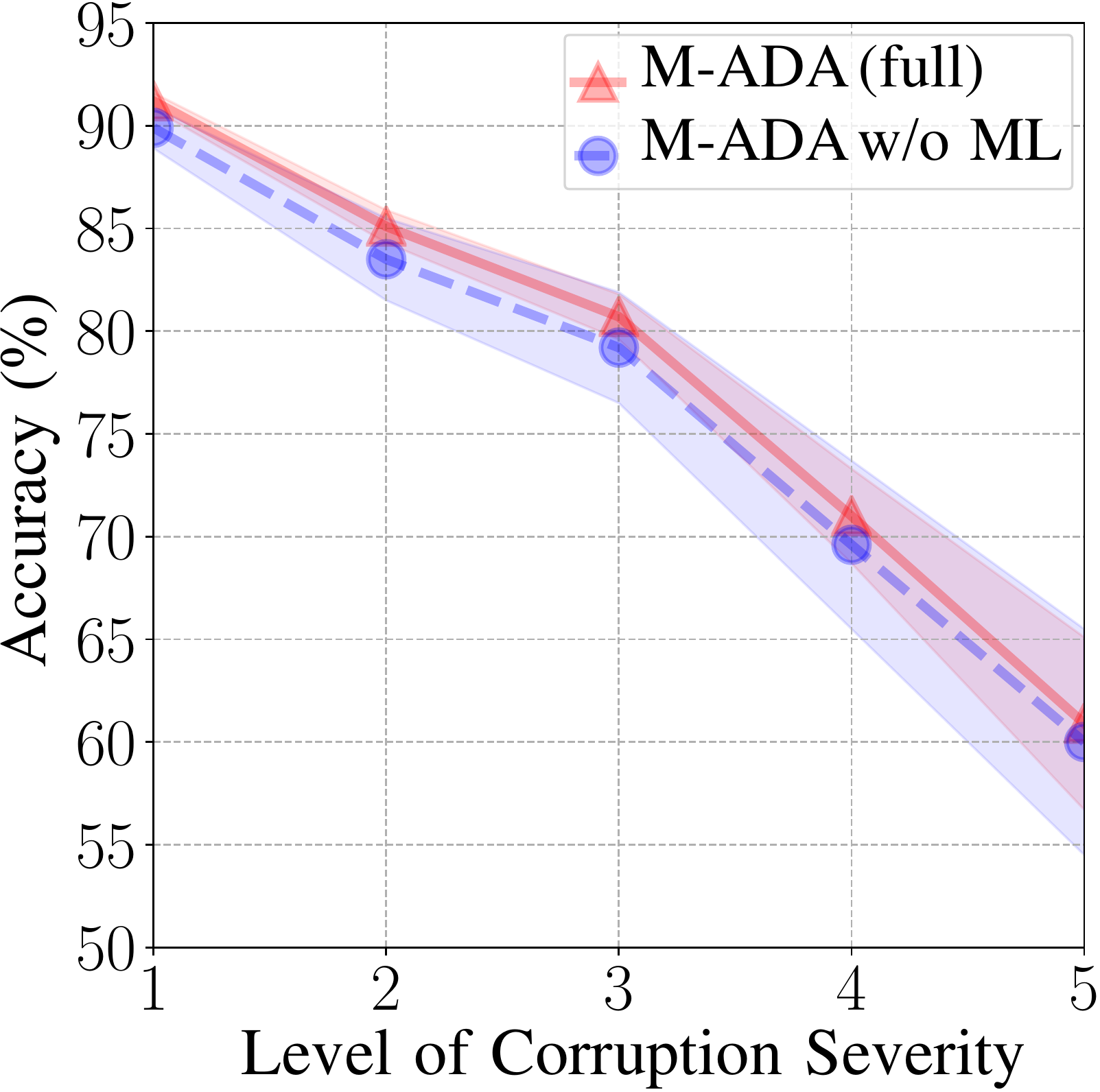}
}
\subfigure{
\includegraphics[width=0.23\linewidth]{impulse_noise.pdf}
}
\subfigure{
\includegraphics[width=0.23\linewidth]{shot_noise.pdf}
}
\end{center}
\vspace{-1em}
\caption{Validation of meta-learning scheme. 
%on CIFAR-10-C \cite{hendrycks2019benchmarking}. 
Five levels of severity are evaluated on each unseen corruption.
\textbf{From left to right:} (a) \textit{Gaussian Noise}; (b) \textit{Speckle Noise} ; (c) \textit{Impulse Noise}; and (d) \textit{Shot Noise}.
}
\label{fig_cifar10meta-supp}
\end{figure*}

{\bf Validation of meta-learning scheme:} The results of four kinds of unseen corruptions are shown in Fig.~\ref{fig_cifar10meta-supp}. As seen, M-ADA can significantly reduce variance and yield better performance across all levels of severity. The experimental results prove that the meta-learning scheme plays a key role to improve the training stability and classification accuracy. This is extremely important when performing adversarial domain augmentation in challenging conditions.

\begin{figure*}[t]
\begin{center}
\subfigure{
	\includegraphics[width=0.3\linewidth]{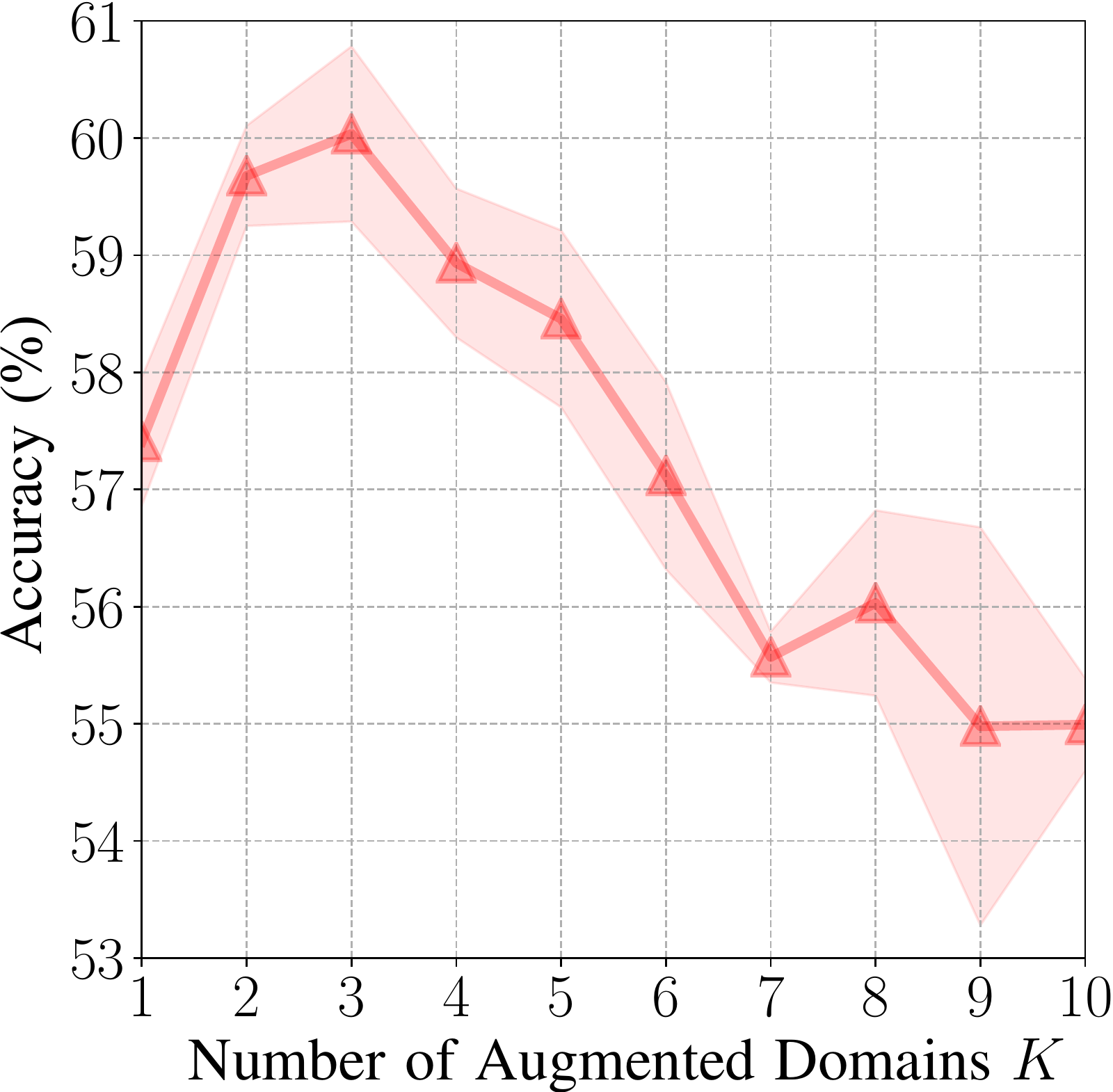}
}
\subfigure{
	\includegraphics[width=0.3\linewidth]{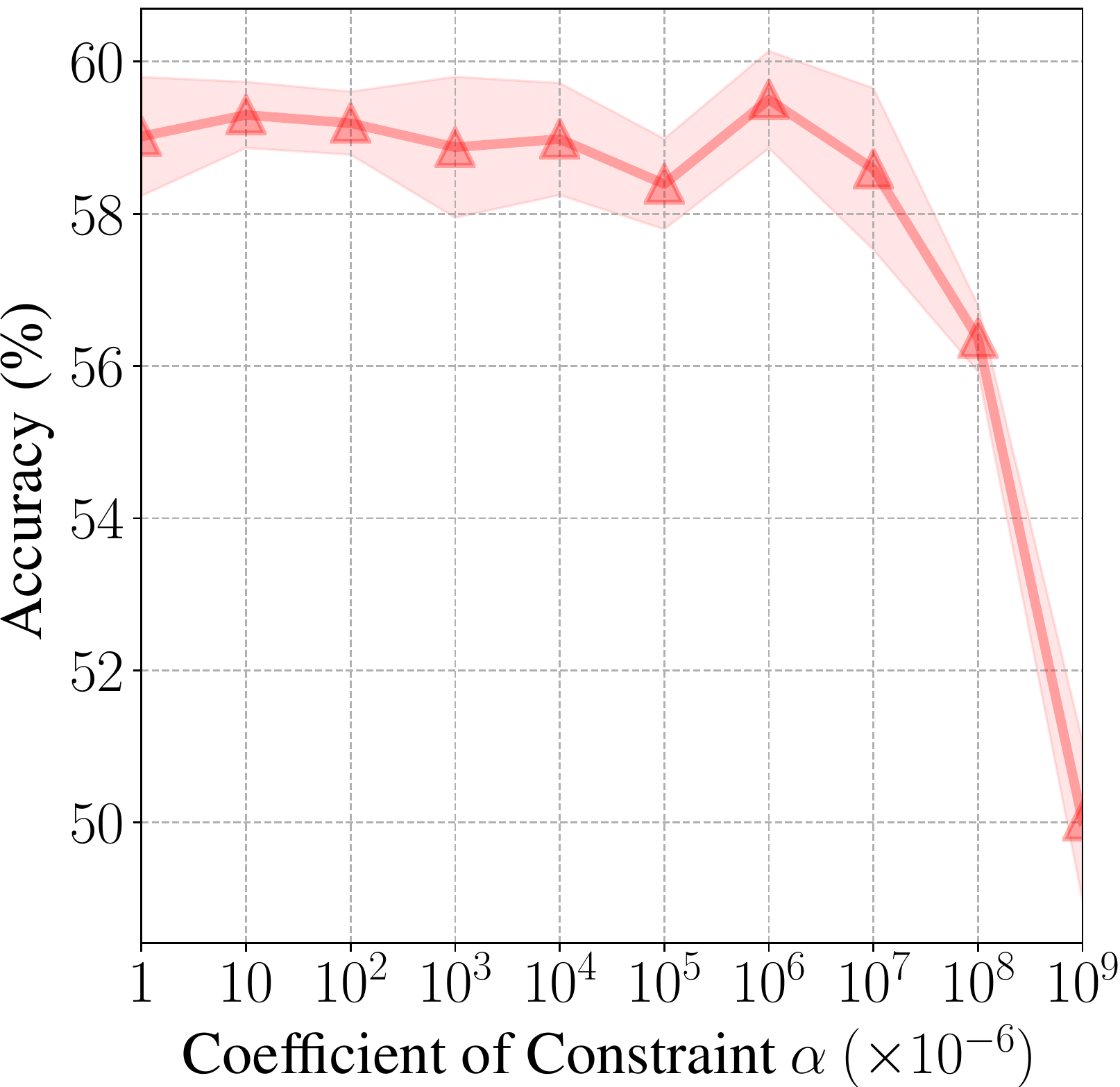}
}
\subfigure{
	\includegraphics[width=0.3\linewidth]{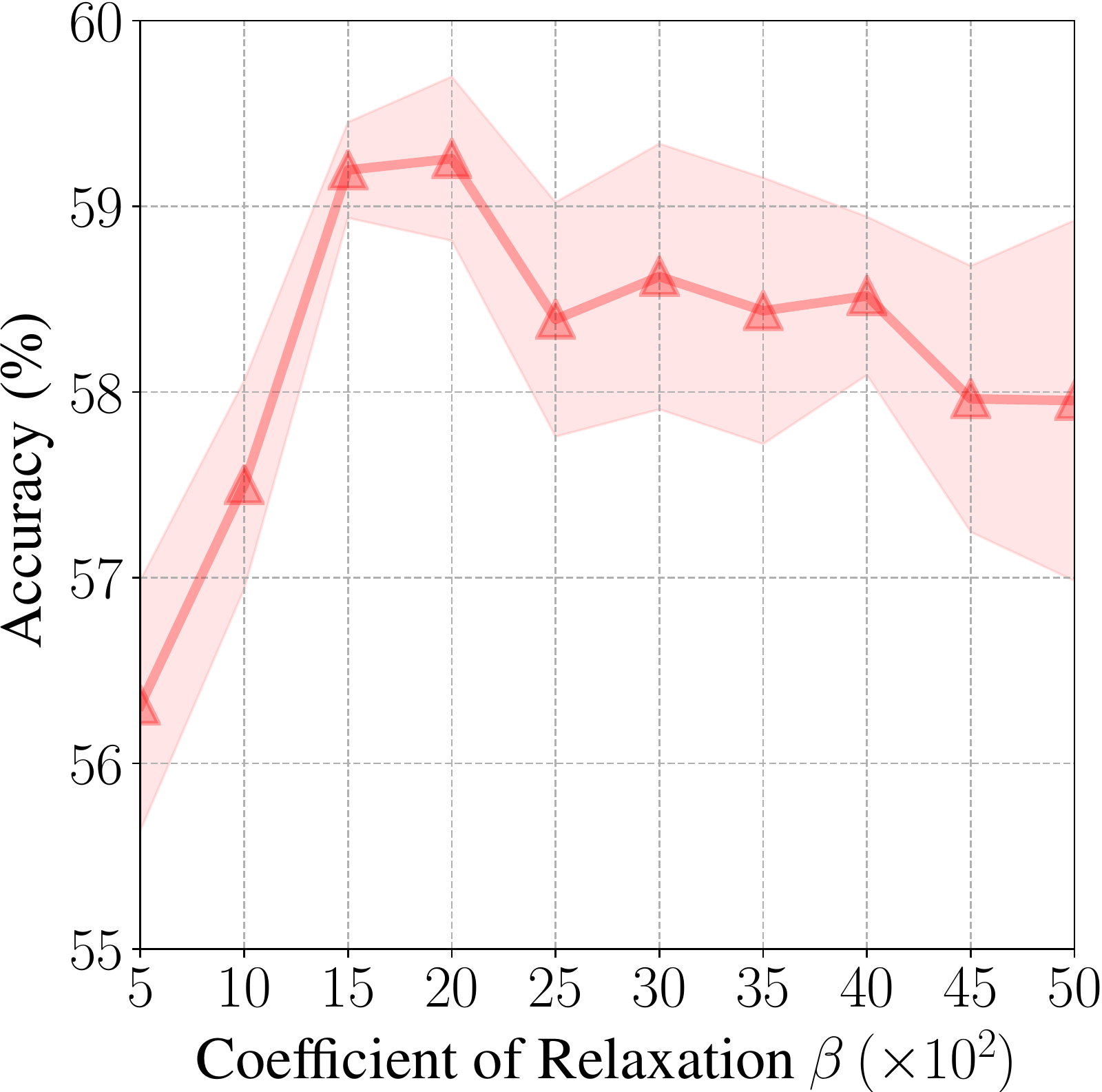}
}
\end{center}
\vspace{-1em}
\caption{Hyper-parameter tuning of $K$, $\alpha$, and $\beta$. We set $K=3$, $\alpha=1.0$, and $\beta=2.0\times10^3$ according to the best accuracy.} 
\label{fig_params-supp}
\end{figure*}

{\bf Hyper-parameter tuning of $K$, $\alpha$, and $\beta$:} We study the effect of three important hyper-parameters of M-ADA: the number of augmented domains ($K$), the distance between the source and augmented domain in the embedding space ($\alpha$), and the deviation between the source and augmented domain ($\beta$). We plot the accuracy curve under different $K$, $\alpha$, and $\beta$ in Fig.~\ref{fig_params-supp}. In Fig.~\ref{fig_params-supp}~(left), we find that the accuracy reaches the summit when $K=3$ and keeps falling with $K$ increasing. This is due to the fact that excessive adversarial samples above a certain threshold will increase the instability and degrade the robustness of the model. Since the distance between the augmented and source domain increases as $K$ increases, a large $K$ may break down the constraint of semantic consistency yielding inferior model training.
In Fig.~\ref{fig_params-supp}~(middle), we find that the accuracy reaches the summit when $\alpha=1.0$ and keeps falling with $\alpha$ increasing. This is because large $\alpha$ will make the source and augmented domain too close in the embedding space, yielding limited domain transportation.
In Fig.~\ref{fig_params-supp}~(right), we observe that the accuracy reaches the summit when $\beta=2.0 \times 10^3$ and drops slightly when $\beta$ increases. This is because large $\beta$ will produce domains too far way from the source $\mathcal{S}$ and even reach out of the manifold in embedding space.

\subsection{Comparison of Different $\mathcal{L}_{\mathrm{relax}}$}

WAEs employ Wasserstein metric to measure the distribution distance between the input and reconstruction, which is desirable for domain augmentation. So the reconstruction error $\mathcal{L}_{\mathrm{relax}}= \Vert \mathbf{x}^+- V(\mathbf{x}^+) \Vert^2$ indicates if $\mathbf{x}^+$ lie in the same distribution as $\mathbf{x}$. Using WAE instead of vanilla AE is the key design to achieve this goal (Tab.~\ref{tbl:R3-supp}). Additionally, our experiments indicate that $\Vert V(\mathbf{x})- V(\mathbf{x}^+) \Vert^2$ has better relaxation effect and yields improved accuracy. The distribution distance is more reliable in the reconstruction space where Wasserstein prior has been applied. 

\begin{table}[h]
% \vspace{-.4em}
\centering
\resizebox{0.85\linewidth}{!}{
\begin{tabular}{@{}l|c|c|c@{}}
\hline
       & $\Vert \mathbf{x} - \mathbf{x}^+ \Vert^2$  & Vanilla AE & WAE\\
\hline
Digits & $55.71\%$  & $58.67\%$ & $59.49\%$\\
CIFAR-10-C & $62.03\%$  & $63.34\%$ & $65.59\%$\\
\hline
\end{tabular}}
\vspace{0.8em}
\caption{Accuracy comparison using different relaxation terms.}
\label{tbl:R3-supp}
\end{table}

\subsection{Comparison on CIFAR-10-C}

We train all models on clean data, {\it i.e.}, CIFAR-10, and test them on corruption data, {\it i.e., CIFAR-10-C}. In this case, there are totally $19$ unseen testing domains. We present the result of each corruption with the highest severity in Tab.~\ref{tab:cifar10_s5-supp}. We observe that M-ADA substantially outperforms other methods on most corruptions. Specially, in several corruptions such as \textit{Frost}, \textit{Glass blur}, \textit{Gaussian blur}, \textit{Pixelate}, and corruptions related with \textit{Noise}, M-ADA outperforms ERM \cite{koltchinskii2011oracle} with more than 10\%. More importantly, M-ADA has the lowest values on mCE and relative mCE, indicating its strong robustness against image corruptions.

\begin{table*}[t]
\begin{center}

\resizebox{1.\linewidth}{!}{
\begin{tabular}{@{}lcccccccccccc@{}}
\toprule
&\multicolumn{3}{c}{Weather}& \multicolumn{5}{c}{Blur}& \multicolumn{4}{c}{Noise}\\
\cmidrule(lr){2-4} \cmidrule(lr){5-9} \cmidrule(lr){10-13}
&Fog& Snow& Frost& Zoom& Defocus& Glass& Gaussian& Motion& Speckle& Shot& Impulse& Gaussian\\
\hline
ERM \cite{koltchinskii2011oracle}&65.92& 74.36& 61.57& 59.97& 53.71& 49.44& 30.74& 63.81& 41.31& 35.41& 25.65& 29.01  \\
CCSA \cite{motiian2017unified}&66.94& 74.55& 61.49& 61.96& 56.11& 48.46& 32.22& \textbf{64.73}& 40.12& 33.79& 24.56& 27.85 \\
d-SNE \cite{xu2019dsne}&65.99& 75.46& 62.25& 58.47& 53.71& 50.48& 33.06& 63.70& 45.30& 39.93& 27.95& 34.02 \\
% JiGen  \cite{carlucci2019jigasaw}&& & & & & & & & & & & & & & \\
GUD \cite{volpi2018generalizing}&68.29& 76.75& 69.94& 62.95& 56.41& 53.45& 38.33& 63.93& 38.45& 36.87& 22.26& 32.43 \\
\hline
M-ADA w/o $\mathcal{L}_{\mathrm{relax}}$ &66.99& 80.09& 74.93& 54.15& 44.67& 60.57& 30.53& 57.06& 59.88& 59.18& 43.46& 55.07 \\
M-ADA w/o ML&67.68& \textbf{80.91}& 76.20& 65.70& 56.87& \textbf{62.14}& 41.20& 63.86& 60.01& 59.63& 40.04& 55.70  \\
{\bf M-ADA} (full)&\textbf{69.36}& 80.59& \textbf{76.66}& \textbf{68.04}& \textbf{61.18}& 61.59& \textbf{47.34}& 64.23& \textbf{60.88}& \textbf{60.58}& \textbf{45.18}& \textbf{56.88}\\
\bottomrule
\end{tabular}}
\resizebox{0.87\linewidth}{!}{
\begin{tabular}{@{}lcccccccccc@{}}
\toprule
&\multicolumn{7}{c}{Digital}& \\
\cmidrule(lr){2-8}
&Jpeg& Pixelate& Spatter& Elastic& Brightness& Saturate& Contrast& Avg.& mCE &RmCE \\
\hline
ERM \cite{koltchinskii2011oracle}&69.90& 41.07& 75.36& 72.40& \textbf{91.25}& 89.09& \textbf{36.87}& 56.15&  1.00&1.00  \\
CCSA \cite{motiian2017unified}&69.68& 40.94& 77.91& 72.36& 91.00& \textbf{89.42}& 35.83& 56.31&   0.99&0.99 \\
d-SNE \cite{xu2019dsne}&70.20& 38.46& 73.40& 73.33& 90.90& 89.27& 36.28& 56.96&  0.99 &1.00 \\
% JiGen  \cite{carlucci2019jigasaw}&& & & & & & & & & & & & & & \\
GUD \cite{volpi2018generalizing}&74.22& \textbf{53.34}& 80.27& 74.64& 89.91& 82.91& 31.55& 58.26&  0.97&0.95 \\
\hline
M-ADA w/o $\mathcal{L}_{\mathrm{relax}}$ &76.45& 53.13& 80.75& 73.85& 90.86& 87.01& 27.83& 61.92&  0.90&0.86 \\
M-ADA w/o ML&\textbf{77.62}& 52.49& \textbf{81.02}& 75.54& 90.69& 86.58& 26.30& 64.22&  0.85 &0.80  \\
{\bf M-ADA} (full)&77.14& 52.25& 80.62& \textbf{75.61}& 90.78& 87.62& 29.71& \textbf{65.59}&  \textbf{0.82}&\textbf{0.77}\\
\bottomrule
\end{tabular}}

\end{center}
\caption{Full version of Tab.~4 in main paper. The models are generalized from clean data to different corruptions. We report the classification accuracy (\%) of 19 corruptions under the corruption level of ``5'' (severest). We also report the mean Corruption Error (mCE) and relative mCE (RmCE) in the last two columns. The lower the better for mCE and RmCE.  
%For each method, we also report mCE and RmCE (Sec.~\ref{sec:setting}) computed over 19 corruptions in the \textbf{last two columns}. The lower the better for mCE and RmCE.
}
\label{tab:cifar10_s5-supp}
\end{table*}

\end{appendices}

\end{document}